\pdfoutput=1
\documentclass[11pt]{article}

\usepackage[final]{acl}

\usepackage{times}
\usepackage{latexsym}

\usepackage[T1]{fontenc}
\usepackage[utf8]{inputenc}

\usepackage{microtype}

\usepackage{inconsolata}

\usepackage{graphicx}

\usepackage{url}            % simple URL typesetting
\usepackage{booktabs}       % professional-quality tables
\usepackage{amsfonts}       % blackboard math symbols
\usepackage{nicefrac}       % compact symbols for 1/2, etc.
\usepackage{xcolor}         % colors
\usepackage{comment}
\usepackage{multirow}
\usepackage{xspace}
\usepackage{subcaption}
\usepackage{pifont}
\usepackage{ragged2e}
\usepackage{enumitem}
\usepackage{longtable}
\usepackage{color, colortbl}

\usepackage{tikz}
\usetikzlibrary{shapes.geometric, arrows.meta, positioning}

\usepackage{float} % to use [H] if needed

\usepackage{textcomp}

\definecolor{Gray}{gray}{0.85}
\definecolor{LightCyan}{rgb}{0.88,1,1}
\newcolumntype{a}{>{\columncolor{LightCyan}}r}

\title{Reasoning Beyond Labels: Measuring LLM Sentiment in Low-Resource, Culturally Nuanced Contexts}

\author{First Author \\
  Affiliation / Address line 1 \\
  Affiliation / Address line 2 \\
  Affiliation / Address line 3 \\
  \texttt{email@domain} \\\And
  Second Author \\
  Affiliation / Address line 1 \\
  Affiliation / Address line 2 \\
  Affiliation / Address line 3 \\
  \texttt{email@domain} \\}

\author{%
 Millicent Ochieng$^{1}$, Anja Thieme$^{1}$, Ignatius Ezeani$^{2}$, Risa Ueno$^{1}$,\\
 \textbf{Samuel Maina$^{1}$, Keshet Ronen$^{3}$, Javier González$^{1}$,   Jacki O’Neill$^{1}$}  \\ \\
\footnotesize
$^1$Microsoft Research, 
$^2$Lancaster University, 
$^3$University of Washington\\
}

\begin{document}
\maketitle
\begin{abstract}

Sentiment analysis in low-resource, culturally nuanced contexts challenges conventional NLP approaches that assume fixed labels and universal affective expressions. We present a diagnostic framework that treats sentiment as a context-dependent, culturally embedded construct, and evaluate how large language models (LLMs) reason about sentiment in informal, code-mixed WhatsApp messages from Nairobi youth health groups. Using a combination of human-annotated data, sentiment-flipped counterfactuals, and rubric-based explanation evaluation, we probe LLM interpretability, robustness, and alignment with human reasoning. Framing our evaluation through a social-science measurement lens, we operationalize and interrogate LLMs outputs as an instrument for measuring the abstract concept of sentiment. Our findings reveal significant variation in model reasoning quality, with top-tier LLMs demonstrating interpretive stability, while open models often falter under ambiguity or sentiment shifts. This work highlights the need for culturally sensitive, reasoning-aware AI evaluation in complex, real-world communication.

\end{abstract}

\section{Introduction}
Sentiment analysis is a prevalent NLP technique used to obtain meaningful information and semantics from text~\cite{onyenwe2020impact}. It is often conflated with emotion detection, which focuses on the psychological state or mood articulated in messages~\cite{nandwani2021review}; or opinion mining such as consumer sentiment~\cite{burnham2024sentiment}. Instead, sentiment analysis primarily determines polarity in the intent behind a written message as positive, negative, or neutral~\cite{nandwani2021review}. 

Speech Act Theory by ~\citet{austin1975things} further differentiates between the form of an utterance (locution); its purpose and effect on the hearer (illocution); and the real-world impact (perlocution). This theory highlights that what a reader understands from a message depends on words choices; their individual meanings; ordering; as well as lexical or syntactic variations. Combined, these introduce significant ambiguity on how inferences about a message are drawn~\cite{corvi2025taxonomizing}. Furthermore, the social semiotics theory by~\citet{halliday2014language} emphasizes that sentiment is not just a linguistic phenomenon, and also deeply embedded in social and cultural contexts; highlighting how emotions are conveyed and interpreted based on cultural norms and values~\cite{ Zhang2024sentiment}. 

In this paper, we acknowledge that interpreting or measuring sentiment can be difficult —particularly in informal, multilingual, under-resourced, and culturally nuanced communication contexts. Expressions of emotion and attitude are shaped by local language practices, shared cultural knowledge, and interactional context \cite{Matsumoto1990,Lindquist2021,Fang2022}. In real-world communications such as youth chat, social media, or hyperlocal exchanges among multilingual speakers, language is frequently code-mixed\footnote{The practice where multilingual speakers fluidly shift between languages in conversation}, fluid, and shaped by the moment—that is, influenced by who is speaking and who is listening, the topic being discussed, the speaker’s emotional tone, or intentions at that time, and the setting (e.g., online chat vs. in-person talk). Meanings are negotiated, implicit, and frequently ambiguous—making sentiment difficult to interpret, even for humans, especially when removed from the original platform or context of exchange \cite{10.1145/958160.958167}. These complexities do not just complicate classification—they challenge the very \textit{measurement} of sentiment. As argued by~\citet{wallach2025position}, evaluating GenAI models requires treating such tasks as a social science measurement challenge, where abstract, culturally-contested concepts must be systematically defined and carefully connected to observable indicators.

 % To support interpretive reliability in this setting, we developed a structured annotation protocol that captures ambiguity, cultural nuance, and context-specific sentiment expression (see Appendix~\ref{appendix:annotation}).

\begin{table*}[h]
\centering
\scriptsize
\begin{tabular}{|m{3cm}|m{6cm}|m{3cm}|m{3cm}|}
\toprule
\textbf{Example} & \textbf{Complexities} & \textbf{Annotator 1} & \textbf{Annotator 2} \\
\midrule

\raisebox{-0.2em}{\includegraphics[height=1em]{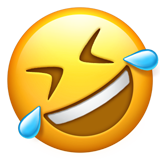}}
\raisebox{-0.2em}{\includegraphics[height=1em]{images/rolling_on_the_floor_laughing.png}}
\raisebox{-0.2em}{\includegraphics[height=1em]{images/rolling_on_the_floor_laughing.png}}\textit{uyu sasa anachoma manzee} \newline
“\raisebox{-0.2em}{\includegraphics[height=1em]{images/rolling_on_the_floor_laughing.png}}
\raisebox{-0.2em}{\includegraphics[height=1em]{images/rolling_on_the_floor_laughing.png}}
\raisebox{-0.2em}{\includegraphics[height=1em]{images/rolling_on_the_floor_laughing.png}} this guy is now messing up, bro” &

\textbf{Shorthand:} ``uyu'' instead of ``huyu'' \newline
\textbf{Urban slang (Sheng):} ``anachoma'', ``manzee'' \newline
\textbf{Tone:} \textit{friendly teasing, mockery, or social critique} \newline
\textbf{Emoji use:} \raisebox{-0.2em}{\includegraphics[height=1em]{images/rolling_on_the_floor_laughing.png}} \newline
\textbf{Code-mixing:} Swahili-Sheng blend \newline
\textbf{Cultural reference:} \textit{Assumes shared understanding of local slang, social behaviors, and norms} &

\textbf{Label:} Negative \newline
\textbf{Notes:} \textit{We see the ridicule and embarrassment from the persona and the audience despite the laugh.} &

\textbf{Label:} Positive \newline
\textbf{Notes:} \textit{Expresses criticism with amusement portrayed with laughing emojis.} \\
\midrule

\textit{Nmeacha izea} \newline “I'm sorry, I have stopped” &
\textbf{Code-mixing:} Swahili-Sheng blend \newline
\textbf{Urban slang (Sheng):} ``izea'' \newline
\textbf{Shorthand:} ``nmeacha'' \newline
\textbf{Tone:} flat or understated \newline
\textbf{Ambiguity:} lacks strong emotional cues &

\textbf{Label:} Neutral \newline
\textbf{Notes:} \textit{We see a casual apology that doesn’t express strong emotion.} &

\textbf{Label:} Positive \newline
\textbf{Notes:} \textit{Speaker is apologetic and remorseful.} \\
\midrule

\textit{U can't see the future but God can} \newline “You can't see the future but God can” &
\textbf{Shorthand:} ``U'' for ``you'' \newline
\textbf{Religious expression:} appeals to divine foresight \newline
\textbf{Tone:} factual or reassuring \newline
\textbf{Cultural context:} common in faith-based communication \newline
\textbf{Ambiguity:} sentiment depends on interpretation of tone/intention &

\textbf{Label:} Neutral \newline
\textbf{Notes:} \textit{A remark without strong personal emotion.} &

\textbf{Label:} Positive \newline
\textbf{Notes:} \textit{Speaker expresses trust in God, offering reassurance.} \\
\midrule

\textit{Hello, guys yani mko tu na mmenyamaza??} \newline “Hello, guys are online and you are quiet?” &
\textbf{Code-mixing:} Swahili-English blend \newline
\textbf{Tone:} questioning, possibly sarcastic \newline
\textbf{Social cue:} expectation of group participation \newline
\textbf{Ambiguity:} tone varies between concern and frustration &

\textbf{Label:} Negative \newline
\textbf{Notes:} \textit{We see disappointment and negative shock from the persona on why people are so quiet.} &

\textbf{Label:} Neutral \newline
\textbf{Notes:} \textit{Expresses concern and curiosity on the silence of the group.} \\
\midrule

\textit{Yes I eat too much iz it normal} \newline “Yes I eat too much is it normal” &
\textbf{Shorthand:} ``iz'' for ``is'', informal tone \newline
\textbf{Self-disclosure:} reveals possible worry \newline
\textbf{Ambiguity:} phrased as a question, unclear tone; genuine concern vs casual comment &

\textbf{Label:} Negative \newline
\textbf{Notes:} \textit{We see worry and distress about too much eating, suggests a negative sentiment.} &

\textbf{Label:} Neutral \newline
\textbf{Notes:} \textit{Question seeking clarification.} \\
\bottomrule
\end{tabular}
\caption{Examples of annotator disagreement illustrating cultural and linguistic complexities.}
\label{tab:annotator-disagreement-complex}
\end{table*}

In our work, we treat \textbf{sentiment} not as a fixed label, but as a context-dependent expression of intent. It may be explicit (e.g., “I’m so angry right now”), but more often in our dataset, it appears through muted cues (e.g., “You're always online”)—subtle, culturally and contextually situated, and open to interpretation. We define \textbf{ambiguity} as cases where the intended sentiment is unclear, underspecified, open to multiple readings, or leads to disagreement even among culturally fluent, context-aware annotators—not because the language is misunderstood, but due to differing interpretations of tone or social context (see Table~\ref{tab:annotator-disagreement-complex}).

We use \textbf{cultural nuance} to describe how language practices, religious or affective expressions, and shared social knowledge shape how sentiment is conveyed and perceived. In our dataset, such nuance is embedded within: practices of \textit{code-mixing} (e.g., “kama hauko school shindaapo!!”)\footnote{Swahili-English: “If you’re not in school, stay there.” While casual, this often conveys a dismissive stance, reflecting cultural norm that link education to intellectual legitimacy.}; \textit{local shorthand} (e.g., mm for mimi)\footnote{\textit{mimi} means “me” in Swahili}; \textit{emoji-only or emoji-enhanced }messages via graphical symbols (e.g.,“~\raisebox{-0.2em}{\includegraphics[height=1em]{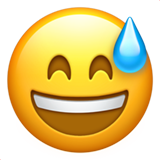}}”) or their textual counterpart the \textit{emoticon} (e.g., ":)")~\cite{ liu2021improving, yoo2021understanding}; \textit{irony}; and \textit{youth-specific slang} (Sheng)\footnote{An urban slang spoken by youth in Kenya, blending Swahili, English, and local languages.}. These elements are often combined to produce rich, but difficult-to-classify, sentiment signals; and these cultural complexities are evident throughout our dataset (see Table~\ref{tab:annotator-disagreement-complex} and Appendix Table~\ref{tab:cultural-nuance})

So far, standard sentiment analysis treats sentiment as a fixed classification problem, assuming a single, context-independent “ground truth” \cite{Mohammad2017,Wankhade2022,Sharma2024}. While LLMs have transformed NLP \cite{Brown2020,Touvron2023,jiang2023mistral7b,OpenAI2023}, their application to sentiment remains narrowly focused on label prediction—rarely leveraging their capacity to explain, paraphrase, or express uncertainty. Conventional evaluations emphasize whether a model assigns the “correct” label, but metrics like accuracy or F1 score obscure how models make decisions and whether those align with human interpretations of sentiment.

We propose a diagnostic approach to sentiment analysis that treats LLMs not only as classifiers but as tools for structuring and probing sentiment in complex, real-world communication. Informed by Wallach et al.'s measurement framework~\cite{wallach2025position} that separates the \textit{conceptualization} of sentiment from its \textit{operationalization}. Our goal is to shift how sentiment is measured in LLMs—from fixed label prediction toward a more interpretive, ambiguity-aware framework. We ask: How do LLMs reason about sentiment in real-world, culturally grounded messages?

To achieve this, we investigate how LLMs reason about sentiment, how they explain their judgments, handle ambiguity, and echo human disagreement. For instance, while a traditional classifier might label “sawa tu~\raisebox{-0.2em}{\includegraphics[height=1em]{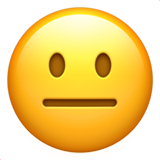}}”\footnote{\textit{sawa tu} means “just okay,” but can imply resignation or frustration depending on tone and context.} as neutral, our framework surfaces the emotional nuance by analyzing emoji, tone, and context, revealing how such utterances can signal quiet frustration or withdrawal. We analyze model explanations, confidence scores, and token-level highlights indicating which parts of the message influenced the model’s judgment, across three evaluation settings: messages with annotator agreement (Gold), disagreement (Ambiguous), and \textbf{sentiment-flipped counterfactuals}\footnote{A \textit{counterfactual} is a sentiment-flipped variant of a real message.} \textbf{(Synthetic)}. These \textit{synthetic} examples are automatically generated by prompting GPT-4 to rewrite real WhatsApp messages in our dataset with their \textit{sentiment flipped (positive to negative or vise versa)}—while preserving meaning, cultural tone and informal language. We guide this process using a structured taxonomy of sentiment-bearing components (e.g., negation, emoji, tone, key phrases; see Appendix~\ref{appendix:transformations}, Table~\ref{tab:transformation-taxonomy}). These counter-factual flips serve as our operationalization of the sentiment concept. Applied in testing whether models respond appropriately to affective changes, we use a dual evaluation protocol with human annotators and LLM-as-judge to assess counterfactual plausibility and explanation quality of the results.

\textbf{This paper makes the following contributions:}
\begin{itemize}
\item We adapt a social-science measurement lens to evaluate model reasoning about language, reframing sentiment analysis as a problem of concept systematization and measurement.
\item We introduce a diagnostic framework to analyze how LLMs reason about sentiment in informal, code-mixed, and culturally embedded communication.
This involves creating synthetic data using a counterfactual approach based on a taxonomy of sentiment components (e.g., negation, emoji, tone).
\item We propose a dual evaluation protocol with human annotators and an LLM-as-judge to assess explanation quality and counterfactual plausibility. Through this, we identify reasoning inconsistencies in LLMs, distinguishing between reducible errors and irreducible ambiguity across evaluation settings.
\end{itemize}

\section{Related Work}

\textbf{Sentiment Analysis in Informal and Multilingual Communication:} While sentiment analysis has largely focused on English-language data from structured text domains like reviews or news, real-world communication, especially in informal, multilingual, and code-mixed contexts presents deeper challenges \cite{article}. Here, sentiment is often more implicit, culturally embedded, and conveyed through tone, slang, or local reference. Prior work on code-mixed sentiment (e.g., Swahili-English, Hindi-English) has highlighted the need for inclusive resources \cite{zhang-etal-2023-multilingual,Doruz2023ASO,Doruz2023RepresentativenessAA,Kaji2023ContextualCS}, but few studies examine low-resource, conversational data in health or community settings. Our work addresses this gap through a WhatsApp corpus of Nairobi youth, where Swahili, English, and Sheng blend in health-related discussions. Rather than focusing solely on prediction accuracy, we use this setting to investigate how LLMs reason about sentiment in complex, everyday language.

\noindent\textbf{Evaluating LLM Reasoning:} Traditional sentiment evaluation relies on metrics like accuracy and F1, which fail to capture how models reason, especially in ambiguous or culturally situated cases \cite{Lyu2024}. To address this, recent work has explored explanation-based evaluation through token attribution, rationales, and confidence scores \cite{joshi2023ertestevaluatingexplanationregularization,dhaini2025evalxnlpframeworkbenchmarkingposthoc}. Other work has shown that LLMs like GPT-4 can serve as evaluators, often approximating human ratings in generation tasks \cite{liu-etal-2023-g}. However, a missing component in this literature is the use of dual evaluation protocols that involve both human and LLM judges applying shared rubrics. Such approaches are particularly valuable in settings with annotator disagreement, where interpretive alignment matters more than single-label accuracy. Our work builds on and extends this direction by systematically comparing model and human evaluations across diverse examples, including ambiguous and counterfactually altered messages.

\noindent\textbf{Counterfactuals and Contrastive Evaluation in NLP:} Counterfactuals offer a powerful tool for probing model reasoning by introducing minimal, targeted changes to input data \cite{yang-etal-2021-exploring}. In sentiment analysis, these typically flip polarity through shifts in tone, negation, or word choice. While prior work often relied on rule-based or synthetic constructions \cite{yang-etal-2021-exploring}, we use GPT-4 to generate sentiment-flipped versions of messages—shifting from positive to negative and vice versa—grounded in a taxonomy of transformation types such as emoji use, phrase substitution, and tone modulation. More broadly, our approach aligns with recent work on problem variation as a diagnostic for reasoning \cite{xu2025re-imagine}, which emphasizes the need for systematic, multi-level perturbations, including counterfactuals to reveal model limitations beyond memorization.

\section{Evaluation as Measurement: Experimental Setup}

\subsection{Dataset and Annotation}

We build on the WhatsApp Chat Dataset originally collected by \citet{Karusala2021ThatCT} and annotated by \citet{mondal-etal-2021-linguistic}, which comprises multilingual conversations among young people living with HIV in informal settlements in Nairobi, Kenya. These discussions, drawn from two health-focused WhatsApp groups moderated by a medical facilitator, are informal, context-rich, and code-mixed across English, Swahili, and Sheng. All messages were anonymized, and ethical protocols from the original collection were strictly followed. The dataset is not publicly released due to sensitivity, but researchers may request access for academic use.

For this study, we developed a structured annotation protocol focused on culturally grounded sentiment, interpretive ambiguity, and context-specific expression. Designed through iterative pilot testing and calibrator discussions (see Appendix~\ref{appendix:annotation}). Two trained annotators — Kenyan youth aged 20–24 — labeled each message for sentiment (positive, negative, neutral), provided English translations where needed, and tagged word-level language identifiers. Messages with annotator disagreement were retained for targeted evaluation. From the full dataset of 6,197 messages, we define three evaluation subsets: the \textbf{Gold Set }(6,121 messages with full annotator agreement), the \textbf{Ambiguous Set }(76 messages with disagreement), and the \textbf{Synthetic Set} (sentiment-flipped messages generated from a pool of 1,547 non-neutral messages using GPT-4), see Table~\ref{tab:sentiment-subset-breakdown}. No post-processing is applied to normalize emojis, punctuation, or shorthand expressions, as these elements are integral to the communicative and emotional tone of the data.

\subsection{Task and Model Setup}

We frame sentiment analysis as a multi-class classification task over informal, multilingual WhatsApp messages. Given an input message, the model is prompted to predict a sentiment label—\textit{positive}, \textit{negative}, or \textit{neutral}—and to generate a natural language explanation (max 200 words). The task is performed via in-context learning using few-shot prompting, with carefully curated examples included in each prompt (see Table \ref{tab:sentiment_prompt} in the Appendix). We evaluate a range of LLMs varying in architecture and size: \texttt{\textbf{GPT-4-Turbo}}, \texttt{\textbf{GPT-4-32k}} \cite{OpenAI2023}, \texttt{\textbf{Gemma-3-27B}} \cite{gemmateam2025gemma3technicalreport}, \texttt{\textbf{LLaMA-3-8B}} \cite{Touvron2023}, \texttt{\textbf{Mistral-7B}} \cite{jiang2023mistral7b}, \texttt{\textbf{OpenChat-3.5}} \cite{wang2024openchatadvancingopensourcelanguage}, and \texttt{\textbf{Phi-4}} \cite{gunasekar2023textbooksneed}. Each model outputs a sentiment label, a natural language explanation, token-level highlights, and a confidence score\footnote{Model-reported confidence where available.} (scaled to 0–5). Evaluation is conducted across three data partitions: Gold, Ambiguous, and Synthetic.

\subsection{Counterfactual Generation Framework}

We use sentiment counterfactuals as a diagnostic tool—aligned with \textit{hypothesis validity} testing \citep{wallach2025position}—to evaluate whether models detect and explain controlled shifts in sentiment. Starting with 1,547 non-neutral messages from the Gold Set, we prompt GPT-4 to generate three sentiment-flipped variants per message, reversing the original sentiment (positive to negative or vice versa) while preserving tone, meaning, and conversational style. We observed that some messages required only minimal lexical substitution, for example, \textit{Napenda wazo lako} (“I like your idea”) → \textit{Sipendi wazo lako} (“I dislike your idea”). However, many messages demanded deeper edits involving tone, intent, or phrasing shifts, such as \textit{Sema tuu niache kukuaibisha} (“Just tell me to stop embarrassing you”) [–]
→ \textit{Sema tu niendelee kukusifu} (“Just say it, so I continue praising you”) [+]. To guide the model in generating high-quality flips across this range of complexity, we developed a taxonomy of sentiment-bearing components—including negation, tone, emoji, and sentiment phrases—which informed the generation prompt (Table~\ref{tab:transformation-taxonomy}). Rather than selecting flips manually, we follow this with a second prompt that asks GPT-4 to select the strongest candidate based on criteria including plausibility, fluency, and contextual fit (Table~\ref{tab:cf_prompts}). This two-step process allows for richer variation at generation time while ensuring higher-quality, interpretable flips. The resulting counterfactuals form the Synthetic Set used in our evaluation. Although this filtering step is not independently validated in our current setup, we discuss its limitations and suggest improvements for future work.

\subsection{Human and LLM-as-Judge Evaluation Protocol}

We evaluate both model explanations and Synthetic data using a structured, rubric-based protocol involving human annotators and GPT-4 as an automated judge. This dual evaluation enables us to assess interpretive quality, generation plausibility, and alignment between human and model judgment. For model explanations, two annotators rated 480 explanations across the Gold (180), Ambiguous (120), and Synthetic (180) sets. All six LLMs under evaluation were included where explanations were available; for the Ambiguous set, only four models (\texttt{LLaMA-3-8B}, \texttt{GPT-4-Turbo}, \texttt{GPT-4-32k}, and \texttt{Gemma-3-27B}) returned usable explanations, with others failing to produce outputs under the same prompt conditions, highlighting the inherent difficulty of these cases. They scored four dimensions—faithfulness, contextual or cultural appropriateness, logical coherence, and clarity/completeness—using a binary (0/1) scale. For the Synthetic data, six annotators evaluated a sample of 50 counterfactual messages on fluency, naturalness, sentiment flip clarity, and meaning preservation, also on a 0/1 scale. GPT-4 was prompted to follow the same rubrics using standardized evaluation instructions (see Tables \ref{tab:explanation_eval_prompt} and \ref{tab:cf_quality_prompt}). This rubric-based protocol represents the \textit{interrogation} step in the measurement framework of \citet{wallach2025position}, where we examine the \textit{content validity} of model explanations (i.e., how well they align with the substance of the sentiment concept) and their \textit{consequential validity} (i.e., how interpretation quality affects downstream understanding and use). Rubric definitions and example annotations are provided in Appendix~\ref{appendix:rubrics}.

\section{Results and Analysis}

\subsection{Overall Model Performance}
We observe that model coverage\footnote{Coverage reflects the percentage of examples for which a model returned a valid sentiment label.} varies substantially across settings, especially under counterfactual perturbation revealing a key axis of performance variation (see Table~\ref{tab:quant-performance}). While top-tier models like \texttt{GPT-4-Turbo} and \texttt{GPT-4-32k}, consistently provide labels for all examples (100\% coverage), several open-weight models—most notably \texttt{LLaMA-3-8B}—show sharp declines, especially in the Synthetic set, where coverage drops as low as 37.6\%. This sharp drop suggests that even fluent, sentiment-flipped rewrites can disrupt model processing, exposing fragility to subtle changes in tone, emoji, or phrasing. 

We further observe that on the Gold Set, all models achieve strong average F1 scores. The best performance is observed from \texttt{GPT-4-32k} (0.90), \texttt{Mistral-7B} (0.90), and \texttt{Gemma-3-27B} (0.89). Most models maintain balance across sentiment classes, but class-specific performance still varies. \texttt{LLaMA-3-8B} underperforms markedly on negative sentiment (0.51), pointing to difficulty detecting more implicit or culturally nuanced negativity. Neutral sentiment is generally the most challenging class, echoing prior findings on underspecified affect and implicit tone. These results establish strong baselines while highlighting gaps in both robustness and class sensitivity that motivate further analysis of model reasoning and explanation quality.

\subsection{Reasoning Quality in LLM Explanations}
\begin{figure}[h]
    \centering
    \includegraphics[width=1\linewidth]{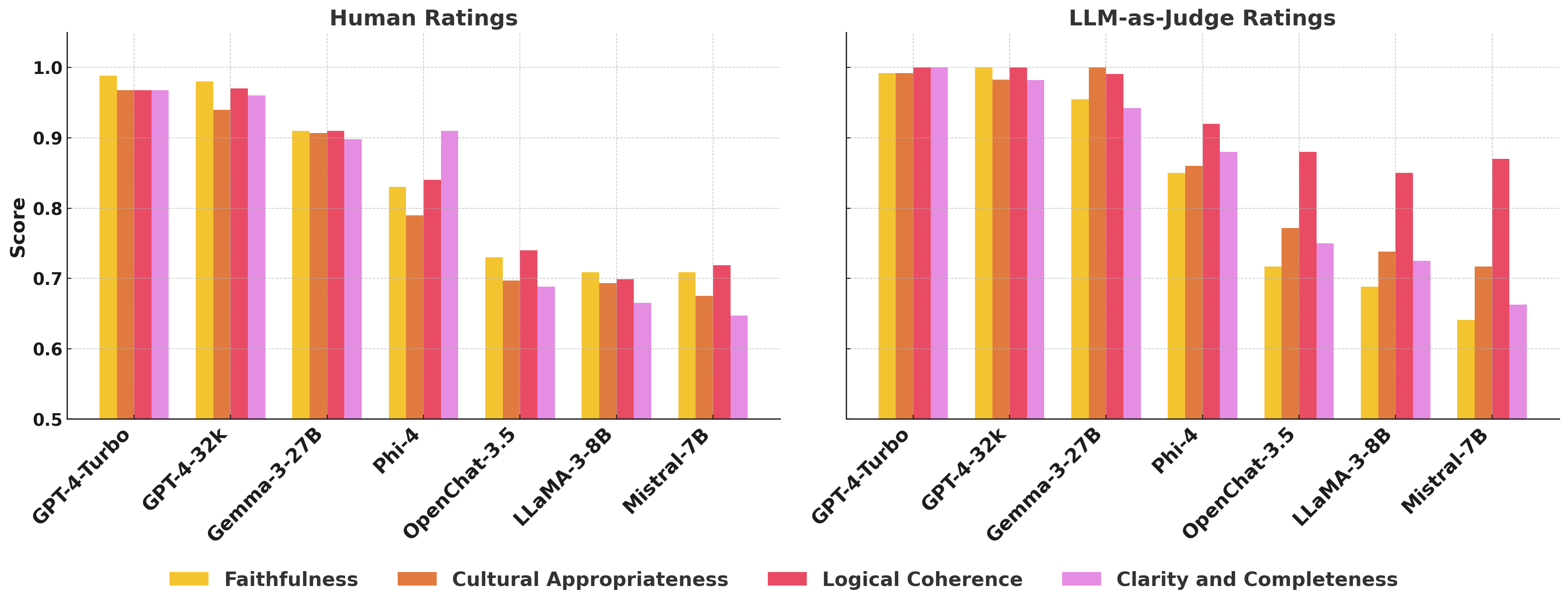}
    \caption{Rubric-based average explanation scores across models.}
    \label{fig:explanation-quality}
\end{figure}
We evaluated explanation quality using rubric-based scores from both human annotators and GPT-4-based LLM-as-judge systems, see Figure~\ref{fig:explanation-quality}. Across all models and dimensions, we observe broad agreement in relative rankings between the two rating sources, though LLM-as-Judge ratings tend to be more generous overall. \texttt{GPT-4-32k} and \texttt{GPT-4-Turbo} consistently achieve top scores across all rubrics, with near-perfect ratings from both humans and LLMs. \texttt{Gemma-3-27B} also performs well, with high ratings for faithfulness, logical coherence and cultural appropriateness, though with modest drops in clarity. By contrast, \texttt{Phi-4}, \texttt{OpenChat-3.5}, \texttt{LLaMA-3-8B}, and \texttt{Mistral-7B} show significantly lower performance, particularly on faithfulness and clarity—dimensions most sensitive to hallucination and underspecification. Human raters were notably stricter in these areas, especially for open-weight models, revealing that LLM-based evaluations may overestimate explanation quality. Despite these differences in score magnitude, the rubric-level trends are consistent: Logical Coherence is the strongest dimension across most models, while Faithfulness, Cultural Appropriateness and Clarity \& Completeness expose key weaknesses in less capable systems. Interestingly, \texttt{Mistral-7B}, which led in classification F1, ranks lowest in explanation quality by both rating sources, highlighting a persistent disconnect between predictive accuracy and reasoning quality. Conversely, the strongest models (\texttt{GPT-4} variants and \texttt{Gemma}) exhibit both high classification performance and robust explanatory reasoning. These findings emphasize the importance of explanation-focused evaluation, as high task accuracy alone may mask serious limitations in model understanding and reasoning.

\subsection{Probing LLM Robustness to Synthetic Set (Counterfactual Flips)}
\begin{table}[H]
\centering
\scriptsize
\begin{tabular}{lcc}
\toprule
\textbf{Criterion} & \textbf{Human} & \textbf{LLM-as-Judge} \\
\midrule
Fluency              & 0.89 & 1.00 \\
Naturalness          & 0.68 & 0.97 \\
Flip Clarity         & 0.79 & 0.98 \\
Meaning Preservation & 0.78 & 0.58 \\
\bottomrule
\end{tabular}
\caption{Average rubric-based scores for synthetic flips.}
\label{tab:synthetic-eval}
\end{table}

We categorized each counterfactual by its main transformation and found that flips most commonly altered sentiment-bearing keywords, phrases, tone, and emoji—components central to both explicit and stylistic sentiment signaling, see Figure~\ref{fig:transformations}. Less frequent were transformations involving negation, intent framing, or valence modulation, which require more interpretive reasoning. From our analysis, GPT-4 often produced plausible synthetic flips (see examples in Table~\ref{tab:synthetic-success}). We assessed the quality of the synthetic flips using rubric-based judgments from both human annotators and LLMs-as-Judges (\texttt{GPT-4-Turbo} and \texttt{GPT-4-32k}), see Table~\ref{tab:synthetic-eval}. LLM ratings were uniformly high—near-perfect in fluency, naturalness, flip clarity, and slightly lower for meaning preservation. Human annotators, however, were notably stricter, especially in naturalness and meaning preservation, revealing significant gaps in how surface-level and semantic quality are perceived. In particular, humans flagged many cases as semantically incorrect or stylistically unnatural, despite their formal fluency. Manual analysis revealed that \textit{positive-to-negative} flips posed greater challenges. LLMs frequently overcorrected, introducing harsh or exaggerated tone, especially in code-mixed inputs (see Table~\ref{tab:synthetic-failures}). Conversely, \textit{negative-to-positive} flips tended to be smoother and more culturally appropriate. While human raters penalized positive-to-negative flips for harsh tone or topic drift, LLMs-as-Judges often gave high marks even in such cases—suggesting they were less sensitive to subtle shifts in meaning or register. While the flipped sentiment was often correct, the model struggled with non-English and code-mixed inputs, frequently normalizing local shorthand, translating content into English, or rewriting messages in Standard Swahili (Kiswahili Sanifu), thereby altering the original language composition (see third example in Table~\ref{tab:synthetic-success}). 
% These findings emphasize that while LLMs are capable of generating fluent and structurally coherent counterfactuals, human-centered evaluation reveals persistent risks in meaning preservation and cultural appropriateness.

\begin{table}[H]
\centering
\scriptsize
\setlength{\tabcolsep}{4pt}
\begin{tabular}{lccc}
\toprule
\textbf{Model} & \textbf{Eff. F1 (Pre-CF)} & \textbf{Eff. F1 (Post-CF)} & \textbf{$\Delta$ Post–Pre} \\
\midrule
\texttt{GPT-4-Turbo}     & 0.960 & 0.980 & +0.020 \\
\texttt{GPT-4-32k}       & 0.970 & 0.980 & +0.010 \\
\texttt{Phi-4}           & 0.940 & 0.786 & –0.154 \\
\texttt{Gemma-3-27B}     & 0.940 & 0.466 & –0.474 \\
\texttt{Mistral-7B}      & 0.892 & 0.466 & –0.425 \\
\texttt{OpenChat-3.5}    & 0.910 & 0.441 & –0.469 \\
\texttt{LLaMA-3-8B}      & 0.783 & 0.349 & –0.434 \\
\bottomrule
\end{tabular}
\caption{Effective F1 before and after counterfactual sentiment flips.}
\label{tab:effective-f1}
\end{table}

To quantify model robustness under transformation, we compute \textit{Effective F1}—the product of F1 and coverage. As shown in Table~\ref{tab:effective-f1}, both \texttt{GPT-4-Turbo} and \texttt{GPT-4-32k} maintained high or improved post-flip performance (up to 0.980). In contrast, mid-sized and open models suffered significant drops (0.40–0.47), driven by both misclassification and partial outputs. Notably, \texttt{Phi-4} preserved coverage but underperformed on positive flips, indicating brittle generalization. Beyond label accuracy, explanation quality further reveals this fragility. On the Synthetic Set, only the GPT-4 variants produced consistently faithful, coherent, culturally grounded, and context-sensitive reasoning. Other models often generated fluent but incorrect explanations after sentiment was flipped, with sharp drops in faithfulness and completeness—especially for \texttt{Mistral-7B}, \texttt{OpenChat-3.5}, and \texttt{LLaMA-3-8B} (Table~\ref{tab:explanation_quality}).

\subsection{How does model confidence and alignment reflect interpretive ambiguity?}
\begin{table}[H]
\centering
\scriptsize
\setlength{\tabcolsep}{3pt}
\begin{tabular}{lccc}
\toprule
\textbf{Model} & \textbf{Avg. Conf.} & \textbf{Coverage (\%)} & \textbf{Eff. Conf.} \\
\midrule
\texttt{GPT-4-Turbo}     & 4.639 & 100.0 & 4.64 \\
\texttt{GPT-4-32k}       & 4.440 & 100.0 & 4.44 \\
\texttt{Phi-4}           & 4.711 & 99.5  & 4.69 \\
\texttt{Gemma-3-27B}     & 4.698 & 47.6  & 2.24 \\
\texttt{OpenChat-3.5}    & 4.249 & 47.4  & 2.01 \\
\texttt{Mistral-7B}      & 4.132 & 47.6  & 1.97 \\
\texttt{LLaMA-3-8B}      & 3.981 & 37.6  & 1.50 \\
\bottomrule
\end{tabular}
\caption{Effective Confidence on the Synthetic Set.}
\label{tab:effective-confidence}
\end{table}
To assess confidence calibration, we report average model confidence and coverage across the Gold and Synthetic Sets (Table~\ref{tab:appendix-confidence-coverage}). While most models maintain high confidence on the Gold Set, only the GPT-4 variants and \texttt{Phi-4} sustain both high confidence and near-complete coverage on counterfactual inputs. In contrast, models like \texttt{Gemma-3-27B} and \texttt{OpenChat-3.5} appear overconfident despite skipping over half of the flipped messages. To quantify this further, we compute an Effective Confidence score (confidence × coverage), reported in Table~\ref{tab:effective-confidence}, revealing a sharp drop for open models—underscoring their brittleness under minimal sentiment shifts. Although the Gold Set contains messages with full human agreement, models show only moderate alignment with one another. As shown in Figure~\ref{fig:model-agreement}, the highest agreement is observed between \texttt{Gemma-3-27B} and \texttt{Phi-4} ($\kappa = 0.73$), and between \texttt{GPT-4-Turbo} and \texttt{GPT-4-32k} ($\kappa = 0.70$). However, other pairings show weaker agreement—such as \texttt{GPT-4-Turbo} and \texttt{Mistral-7B} ($\kappa = 0.48$)—despite similar average F1 scores. This suggests that even on “clear” cases, LLMs diverge in interpretation, reflecting differences in how they weigh tone, cues, and cultural context.

\section{Discussion}

\paragraph{LLMs-as-Generators: Crafting Cultural Counterfactuals}
Using GPT-4 to generate sentiment-flipped counterfactuals revealed both the model’s strengths and its limitations. Often, it produced fluent, contextually appropriate flips that successfully reversed sentiment while preserving tone and informal style. However, our diagnostic analysis surfaced key weaknesses. Flips from positive to negative frequently introduced exaggerated emotional intensity, suggesting the model struggles to calibrate negative sentiment in subtle, conversational contexts. Additionally, while GPT-4 provided self-reported labels for the components it modified (e.g., tone, emoji, phrasing), these attributions were often imprecise or inconsistent. These findings underscore both the potential and fragility of using LLMs to generate culturally grounded synthetic data—and highlight the continued need for more iteration in prompt instructions as well as human oversight when precision over tone, meaning, and linguistic structure is essential.

\paragraph{LLMs-as-Judges: Evaluating Counterfactuals}
We used GPT-4 as a judge to assess the quality of sentiment-flipped messages—selecting the best rewrite among three generated variants and then scoring the selected flip for fluency, naturalness, meaning preservation, and successful sentiment reversal. This approach streamlined evaluation and scaled the generation pipeline. In many cases, GPT-4's selections aligned with the human judgments. However, because these decisions rely entirely on the model’s internal criteria, we observed inconsistencies—especially for non-English messages with culturally layered meaning. For instance, some selected flips introduced subtle shifts in tone or more formal phrasing, reducing cultural fidelity even when sentiment was accurately reversed. In other cases, we observe that GPT-4 successfully produced plausible flips that changed a message’s perceived sentiment, this was achieved in different ways, which do not necessarily reflect the most \textit{minimal} changes to achieve that effect. For example, flipping “Hahaha” (+) could be achieved by “Not funny” (-) or “Ughhh,” (-) or “This is not funny at all” (-). These insights suggests that additional checks should be put into place already at the filtering step to assess if flips are indeed consistent with the tone, phrasing, language composition or cultural meaning of the original message to ensure chosen variants are truly the most faithful transformations. These findings point to the need for human-in-the-loop validation at each stage—particularly when using LLMs to adjudicate nuanced, multilingual language in low-resource settings.
% REVIEW AFTER HUMAN EVAL

\paragraph{Prediction is not `understanding'}
Models such as \texttt{Mistral-7B}, \texttt{Phi-4}, and \texttt{OpenChat-3.5} score competitively on standard metrics, yet generate explanations that often lack coherence, faithfulness, or cultural grounding—especially in cases where sentiment is subtle, indirect, or stylistically embedded, as revealed by human evaluation. These reasoning gaps become even more pronounced under sentiment counterfactuals, with flipped affect lead to sharp performance drops—up to 0.47 F1 for open-weight models—exposing brittle generalization to plausible shifts in tone, emoji, or phrasing. In contrast, \texttt{GPT-4-Turbo} and \texttt{GPT-4-32k} demonstrate greater robustness in both prediction and reasoning, suggesting that scale and stronger instruction tuning support more stable reasoning.

\paragraph{LLMs vary not just in accuracy, but in worldview}
Agreement scores between models remain low, even on the Gold Set, where human annotators were unanimous. This divergence reflects not just model sensitivity to surface cues, but deeper differences in how LLMs encode sentiment priors, cultural nuance, and conversational style. That \texttt{GPT-4-Turbo} and \texttt{Mistral-7B} can yield similar F1 yet diverge in label agreement ($\kappa = 0.48$) illustrates that we are not simply comparing better vs. worse models, but different interpretive frameworks. However, we do not understand the models underlying interpretive frameworks, and how well it maps to existing theory, and consistency in reasoning varies significantly across models, especially open-weight models.  

\paragraph{Confidence is not calibration} While average confidence scores remain high across models, only OpenAI’s models (\texttt{GPT-4-Turbo} and \texttt{GPT-4-32k}) consistently maintain high confidence, full coverage on perturbed data, accurate predictions, and reliable reasoning. In contrast, models such as \texttt{Phi-4} also exhibits high confidence and broad coverage, but manual inspection reveals frequent reasoning errors, highlighting a gap between confidence and correctness.

\paragraph{Sentiment is structured by context}
Our work challenges simplified views of sentiment as binary or fixed, framing it instead as context-dependent and semantically layered. While our initial definition in the annotation protocol and component taxonomy aimed to capture more nuance, more specification is needed. For example, \textbf{\textit{context-dependency}} emerged as central to interpretation, as seen in our annotation examples. There are many aspects that can shape what context-dependency as an element of sentiment means. As illustrated through our study, context can be informed by: the conversation topic (e.g., health advice); cultural norms (e.g., in Kenya); or religious cues; as well as other interpersonal dynamics (e.g., what the recipient of a message assumes or knows about its writer) that can be harder to capture or specify. Yet, future work will need to expand efforts to further systematize and formalize those components of sentiment to be able to achieve more robust evaluation approaches. 

\paragraph{Annotation as a site of interpretive complexity}
Our study highlights the complexities of designing robust annotation protocols for nuanced, real-world data. Annotators frequently encountered edge cases that exposed ambiguity in how sentiment should be labeled, especially when affect was culturally or contextually embedded. This reinforces growing recognition in human-centered NLP that annotation is an interpretive process requiring iteration, theoretical grounding, and thoughtful handling of disagreement.

\section{Conclusion}

We reframe sentiment analysis in low-resource, culturally nuanced contexts as a problem of reasoning, not just classification. Using a diagnostic framework grounded in social-science measurement, we evaluate how LLMs interpret sentiment in multilingual, code-mixed WhatsApp messages from Nairobi youth health groups. Our findings reveal that while top-tier LLMs demonstrate interpretive robustness, open models often fail under ambiguity and cultural nuance, highlighting deep gaps in reasoning quality. As sentiment increasingly becomes a benchmark task for real-world NLP, our work urges a shift from fixed-label accuracy to context-aware, culturally grounded evaluation. Future sentiment systems must be judged not only by what label they assign, but how and why they reason that way.

\section*{Limitations}

While our diagnostic framework offers a deeper lens into sentiment reasoning, several limitations remain: 

(1) Sentiment itself remains an inherently subjective construct. Our LLM-guided systematization of text components like negation, tone, emojis, keywords and phrase rewordings look reasonable (face validity) and may capture the most salient aspects of the sentiment concept (content validity). However, further research is needed to inspect whether this systematization fully specifies all observable criteria connected to sentiment (substantive validity) \cite{wallach2025position}; as well as how the components may relate to one-another; and whether its operationalization via LLM-as-judge is consistent and coherent with the LLMs internal interpretation of these components. 

(2) Our counterfactual generation pipeline uses a two-stage prompting process: GPT-4 first generates three flipped variants of a message, then selects the most plausible one for inclusion. While this filtering step improves fluency and contextual fit, it relies entirely on the model’s internal criteria, which we do not independently validate. Future work should investigate how this selection process affects flip quality, what may be lost or altered during filtering, and incorporate human-in-the-loop checks to ensure that selected flips accurately reflect the intended sentiment transformation and preserve linguistic and contextual fidelity. 

(3) Our analysis focuses on a single dataset—health-related WhatsApp messages from Nairobi youth—which, while rich in cultural nuance, limits generalizability to other sociolinguistic settings. 

\section*{Ethical Consideration}

This study uses anonymized WhatsApp messages from Nairobi youth health groups, collected with consent under prior research protocols. All data were reviewed to remove identifying information and sensitive content. Our use of LLMs to generate synthetic sentiment data in a code-mixed, culturally grounded setting raises important ethical considerations. Language reflects identity, and synthetic rewrites, especially in informal, multilingual contexts must be handled with care to avoid erasing nuance or reinforcing stereotypes. While we designed prompts to preserve tone and intent, LLMs may still encode harmful biases. We emphasize the importance of cultural sensitivity, context-aware evaluation, and collaboration with local experts to ensure respectful and responsible analysis.

% to remove this section while sending to conference
\section*{Acknowledgment}
We thank all annotators and collaborators for their thoughtful contributions to human annotation, and the assessment of language model explanations and synthetic data. Special thanks to Chris Peter Aloo and Clarence Ajevi for their extensive effort in labeling the full dataset, and to Timothy Wamalwa, Abigail Wachira, Faith Ngetich, Peter Kitonyi, and Lorna Muchangi for their contributions to the evaluation process.

\bibliography{custom}

\begin{thebibliography}{37}
\providecommand{\natexlab}[1]{#1}

\bibitem[{Austin(1975)}]{austin1975things}
John~Langshaw Austin. 1975.
\newblock \emph{How to do things with words}.
\newblock Harvard university press.

\bibitem[{Brown et~al.(2020)Brown, Mann, Ryder, Subbiah, Kaplan, Dhariwal, Neelakantan, Shyam, Sastry, Askell, Agarwal, Herbert-Voss, Krueger, Henighan, Child, Ramesh, Ziegler, Wu, Winter, Hesse, Chen, Sigler, Litwin, Gray, Chess, Clark, Berner, McCandlish, Radford, Sutskever, and Amodei}]{Brown2020}
Tom~B. Brown, Benjamin Mann, Nick Ryder, Melanie Subbiah, Jared Kaplan, Prafulla Dhariwal, Arvind Neelakantan, Pranav Shyam, Girish Sastry, Amanda Askell, Sandhini Agarwal, Ariel Herbert-Voss, Gretchen Krueger, Tom Henighan, Rewon Child, Aditya Ramesh, Daniel~M. Ziegler, Jeffrey Wu, Clemens Winter, Christopher Hesse, Mark Chen, Eric Sigler, Mateusz Litwin, Scott Gray, Benjamin Chess, Jack Clark, Christopher Berner, Sam McCandlish, Alec Radford, Ilya Sutskever, and Dario Amodei. 2020.
\newblock \href {https://arxiv.org/abs/2005.14165v4} {Language models are few-shot learners}.
\newblock \emph{Advances in Neural Information Processing Systems}, 2020-December.

\bibitem[{Burnham(2024)}]{burnham2024sentiment}
Michael Burnham. 2024.
\newblock What is sentiment meant to mean to language models?
\newblock \emph{Research \& Politics}, 11(4):20531680241307941.

\bibitem[{Choudhary et~al.(2018)Choudhary, Singh, Bindlish, and Shrivastava}]{article}
Nurendra Choudhary, Rajat Singh, Ishita Bindlish, and Manish Shrivastava. 2018.
\newblock \href {https://doi.org/10.48550/arXiv.1804.00806} {Sentiment analysis of code-mixed languages leveraging resource rich languages}.

\bibitem[{Corvi et~al.(2025)Corvi, Washington, Reed, Atalla, Chouldechova, Dow, Garcia-Gathright, Pangakis, Sheng, Vann et~al.}]{corvi2025taxonomizing}
Emily Corvi, Hannah Washington, Stefanie Reed, Chad Atalla, Alexandra Chouldechova, P~Alex Dow, Jean Garcia-Gathright, Nicholas Pangakis, Emily Sheng, Dan Vann, et~al. 2025.
\newblock Taxonomizing representational harms using speech act theory.
\newblock \emph{arXiv preprint arXiv:2504.00928}.

\bibitem[{Dhaini et~al.(2025)Dhaini, Hussain, Zaradoukas, and Kasneci}]{dhaini2025evalxnlpframeworkbenchmarkingposthoc}
Mahdi Dhaini, Kafaite~Zahra Hussain, Efstratios Zaradoukas, and Gjergji Kasneci. 2025.
\newblock \href {https://arxiv.org/abs/2505.01238} {Evalxnlp: A framework for benchmarking post-hoc explainability methods on nlp models}.
\newblock \emph{Preprint}, arXiv:2505.01238.

\bibitem[{Doğru{\"o}z et~al.(2023{\natexlab{a}})Doğru{\"o}z, Sitaram, Bullock, and Toribio}]{Doruz2023ASO}
A.~Seza Doğru{\"o}z, Sunayana Sitaram, Barbara~E. Bullock, and Almeida~Jacqueline Toribio. 2023{\natexlab{a}}.
\newblock \href {https://api.semanticscholar.org/CorpusID:236460241} {A survey of code-switching: Linguistic and social perspectives for language technologies}.
\newblock In \emph{Annual Meeting of the Association for Computational Linguistics}.

\bibitem[{Doğru{\"o}z et~al.(2023{\natexlab{b}})Doğru{\"o}z, Sitaram, and Yong}]{Doruz2023RepresentativenessAA}
A.~Seza Doğru{\"o}z, Sunayana Sitaram, and Zheng-Xin Yong. 2023{\natexlab{b}}.
\newblock \href {https://api.semanticscholar.org/CorpusID:264806890} {Representativeness as a forgotten lesson for multilingual and code-switched data collection and preparation}.
\newblock In \emph{Conference on Empirical Methods in Natural Language Processing}.

\bibitem[{Fang et~al.(2022)Fang, Rychlowska, and Lange}]{Fang2022}
Xia Fang, Magdalena Rychlowska, and Jens Lange. 2022.
\newblock \href {https://doi.org/10.1007/S41809-022-00102-2/METRICS} {Cross-cultural and inter-group research on emotion perception}.
\newblock \emph{Journal of Cultural Cognitive Science}, 6:1--7.

\bibitem[{Gunasekar et~al.(2023)Gunasekar, Zhang, Aneja, Mendes, Giorno, Gopi, Javaheripi, Kauffmann, de~Rosa, Saarikivi, Salim, Shah, Behl, Wang, Bubeck, Eldan, Kalai, Lee, and Li}]{gunasekar2023textbooksneed}
Suriya Gunasekar, Yi~Zhang, Jyoti Aneja, Caio César~Teodoro Mendes, Allie~Del Giorno, Sivakanth Gopi, Mojan Javaheripi, Piero Kauffmann, Gustavo de~Rosa, Olli Saarikivi, Adil Salim, Shital Shah, Harkirat~Singh Behl, Xin Wang, Sébastien Bubeck, Ronen Eldan, Adam~Tauman Kalai, Yin~Tat Lee, and Yuanzhi Li. 2023.
\newblock \href {https://arxiv.org/abs/2306.11644} {Textbooks are all you need}.
\newblock \emph{Preprint}, arXiv:2306.11644.

\bibitem[{Halliday(2014)}]{halliday2014language}
Michael Alexander~Kirkwood Halliday. 2014.
\newblock Language as social semiotic.
\newblock \emph{The Discourse Studies Reader. Amsterdam: John Benjamins}, pages 263--272.

\bibitem[{Jiang et~al.(2023)Jiang, Sablayrolles, Mensch, Bamford, Chaplot, de~las Casas, Bressand, Lengyel, Lample, Saulnier, Lavaud, Lachaux, Stock, Scao, Lavril, Wang, Lacroix, and Sayed}]{jiang2023mistral7b}
Albert~Q. Jiang, Alexandre Sablayrolles, Arthur Mensch, Chris Bamford, Devendra~Singh Chaplot, Diego de~las Casas, Florian Bressand, Gianna Lengyel, Guillaume Lample, Lucile Saulnier, Lélio~Renard Lavaud, Marie-Anne Lachaux, Pierre Stock, Teven~Le Scao, Thibaut Lavril, Thomas Wang, Timothée Lacroix, and William~El Sayed. 2023.
\newblock \href {https://arxiv.org/abs/2310.06825} {Mistral 7b}.
\newblock \emph{Preprint}, arXiv:2310.06825.

\bibitem[{Joshi et~al.(2023)Joshi, Chan, Liu, Nie, Sanjabi, Firooz, and Ren}]{joshi2023ertestevaluatingexplanationregularization}
Brihi Joshi, Aaron Chan, Ziyi Liu, Shaoliang Nie, Maziar Sanjabi, Hamed Firooz, and Xiang Ren. 2023.
\newblock \href {https://arxiv.org/abs/2205.12542} {Er-test: Evaluating explanation regularization methods for language models}.
\newblock \emph{Preprint}, arXiv:2205.12542.

\bibitem[{Kaji and Shah(2023)}]{Kaji2023ContextualCS}
Arshad Kaji and Manan Shah. 2023.
\newblock \href {https://api.semanticscholar.org/CorpusID:266374893} {Contextual code switching for machine translation using language models}.

\bibitem[{Karusala et~al.(2021)Karusala, Seeh, Mugo, Guthrie, Moreno, John-Stewart, Inwani, Anderson, and Ronen}]{Karusala2021ThatCT}
Naveena Karusala, David~Odhiambo Seeh, Cyrus Mugo, Brandon~L Guthrie, Megan~Andreas Moreno, Grace~C John-Stewart, Irene Inwani, Richard~J. Anderson, and Keshet Ronen. 2021.
\newblock \href {https://api.semanticscholar.org/CorpusID:231844130} {“that courage to encourage”: Participation and aspirations in chat-based peer support for youth living with hiv}.
\newblock \emph{Proceedings of the 2021 CHI Conference on Human Factors in Computing Systems}.

\bibitem[{Lindquist(2021)}]{Lindquist2021}
Kristen~A. Lindquist. 2021.
\newblock \href {https://doi.org/10.1007/S42761-021-00049-7/METRICS} {Language and emotion: Introduction to the special issue}.
\newblock \emph{Affective Science}, 2:91--98.

\bibitem[{Liu et~al.(2021)Liu, Fang, Lin, Cai, Tan, Liu, and Lu}]{liu2021improving}
Chuchu Liu, Fan Fang, Xu~Lin, Tie Cai, Xu~Tan, Jianguo Liu, and Xin Lu. 2021.
\newblock Improving sentiment analysis accuracy with emoji embedding.
\newblock \emph{Journal of Safety Science and Resilience}, 2(4):246--252.

\bibitem[{Liu et~al.(2023)Liu, Iter, Xu, Wang, Xu, and Zhu}]{liu-etal-2023-g}
Yang Liu, Dan Iter, Yichong Xu, Shuohang Wang, Ruochen Xu, and Chenguang Zhu. 2023.
\newblock \href {https://doi.org/10.18653/v1/2023.emnlp-main.153} {{G}-eval: {NLG} evaluation using gpt-4 with better human alignment}.
\newblock In \emph{Proceedings of the 2023 Conference on Empirical Methods in Natural Language Processing}, pages 2511--2522, Singapore. Association for Computational Linguistics.

\bibitem[{Lyu et~al.(2024)Lyu, Apidianaki, and Callison-Burch}]{Lyu2024}
Qing Lyu, Marianna Apidianaki, and Chris Callison-Burch. 2024.
\newblock \href {https://doi.org/10.1162/COLI_A_00511} {Towards faithful model explanation in nlp: A survey}.
\newblock \emph{Computational Linguistics}, 50:657--723.

\bibitem[{Matsumoto(1990)}]{Matsumoto1990}
David Matsumoto. 1990.
\newblock \href {https://doi.org/10.1007/BF00995569/METRICS} {Cultural similarities and differences in display rules}.
\newblock \emph{Motivation and Emotion}, 14:195--214.

\bibitem[{Mohammad(2017)}]{Mohammad2017}
Saif~M. Mohammad. 2017.
\newblock \href {https://doi.org/10.1007/978-3-319-55394-8_4} {Challenges in sentiment analysis}.
\newblock pages 61--83.

\bibitem[{Mondal et~al.(2021)Mondal, Bali, Jain, Choudhury, Sharma, Gitau, O{'}Neill, Awori, and Gitau}]{mondal-etal-2021-linguistic}
Ishani Mondal, Kalika Bali, Mohit Jain, Monojit Choudhury, Ashish Sharma, Evans Gitau, Jacki O{'}Neill, Kagonya Awori, and Sarah Gitau. 2021.
\newblock \href {https://doi.org/10.18653/v1/2021.law-1.7} {A linguistic annotation framework to study interactions in multilingual healthcare conversational forums}.
\newblock In \emph{Proceedings of the Joint 15th Linguistic Annotation Workshop (LAW) and 3rd Designing Meaning Representations (DMR) Workshop}, pages 66--77, Punta Cana, Dominican Republic. Association for Computational Linguistics.

\bibitem[{Nandwani and Verma(2021)}]{nandwani2021review}
Pansy Nandwani and Rupali Verma. 2021.
\newblock A review on sentiment analysis and emotion detection from text.
\newblock \emph{Social network analysis and mining}, 11(1):81.

\bibitem[{O'Neill and Martin(2003)}]{10.1145/958160.958167}
Jacki O'Neill and David Martin. 2003.
\newblock \href {https://doi.org/10.1145/958160.958167} {Text chat in action}.
\newblock In \emph{Proceedings of the 2003 ACM International Conference on Supporting Group Work}, GROUP '03, page 40–49, New York, NY, USA. Association for Computing Machinery.

\bibitem[{Onyenwe et~al.(2020)Onyenwe, Nwagbo, Mbeledogu, and Onyedinma}]{onyenwe2020impact}
Ikechukwu Onyenwe, Samuel Nwagbo, Njideka Mbeledogu, and Ebele Onyedinma. 2020.
\newblock The impact of political party/candidate on the election results from a sentiment analysis perspective using\# anambradecides2017 tweets.
\newblock \emph{Social Network Analysis and Mining}, 10(1):55.

\bibitem[{OpenAI(2023)}]{OpenAI2023}
OpenAI. 2023.
\newblock \href {https://arxiv.org/abs/2303.08774v3} {Gpt-4 technical report}.

\bibitem[{Sharma et~al.(2024)Sharma, Ali, and Kabir}]{Sharma2024}
Neeraj~Anand Sharma, A.~B.M.Shawkat Ali, and Muhammad~Ashad Kabir. 2024.
\newblock \href {https://doi.org/10.1007/S41060-024-00594-X} {A review of sentiment analysis: tasks, applications, and deep learning techniques}.
\newblock \emph{International Journal of Data Science and Analytics 2024 19:3}, 19:351--388.

\bibitem[{Team et~al.(2025)Team, Kamath, Ferret, Pathak, Vieillard, Merhej, Perrin, Matejovicova, Ramé, Rivière, Rouillard, Mesnard, Cideron, bastien Grill, Ramos, Yvinec, Casbon, Pot, Penchev, Liu, Visin, Kenealy, Beyer, Zhai, Tsitsulin, Busa-Fekete, Feng, Sachdeva, Coleman, Gao, Mustafa, Barr, Parisotto, Tian, Eyal, Cherry, Peter, Sinopalnikov, Bhupatiraju, Agarwal, Kazemi, Malkin, Kumar, Vilar, Brusilovsky, Luo, Steiner, Friesen, Sharma, Sharma, Gilady, Goedeckemeyer, Saade, Feng, Kolesnikov, Bendebury, Abdagic, Vadi, György, Pinto, Das, Bapna, Miech, Yang, Paterson, Shenoy, Chakrabarti, Piot, Wu, Shahriari, Petrini, Chen, Lan, Choquette-Choo, Carey, Brick, Deutsch, Eisenbud, Cattle, Cheng, Paparas, Sreepathihalli, Reid, Tran, Zelle, Noland, Huizenga, Kharitonov, Liu, Amirkhanyan, Cameron, Hashemi, Klimczak-Plucińska, Singh, Mehta, Lehri, Hazimeh, Ballantyne, Szpektor, Nardini, Pouget-Abadie, Chan, Stanton, Wieting, Lai, Orbay, Fernandez, Newlan, yeong Ji, Singh, Black, Yu, Hui, Vodrahalli, Greff, Qiu,
  Valentine, Coelho, Ritter, Hoffman, Watson, Chaturvedi, Moynihan, Ma, Babar, Noy, Byrd, Roy, Momchev, Chauhan, Sachdeva, Bunyan, Botarda, Caron, Rubenstein, Culliton, Schmid, Sessa, Xu, Stanczyk, Tafti, Shivanna, Wu, Pan, Rokni, Willoughby, Vallu, Mullins, Jerome, Smoot, Girgin, Iqbal, Reddy, Sheth, Põder, Bhatnagar, Panyam, Eiger, Zhang, Liu, Yacovone, Liechty, Kalra, Evci, Misra, Roseberry, Feinberg, Kolesnikov, Han, Kwon, Chen, Chow, Zhu, Wei, Egyed, Cotruta, Giang, Kirk, Rao, Black, Babar, Lo, Moreira, Martins, Sanseviero, Gonzalez, Gleicher, Warkentin, Mirrokni, Senter, Collins, Barral, Ghahramani, Hadsell, Matias, Sculley, Petrov, Fiedel, Shazeer, Vinyals, Dean, Hassabis, Kavukcuoglu, Farabet, Buchatskaya, Alayrac, Anil, Dmitry, Lepikhin, Borgeaud, Bachem, Joulin, Andreev, Hardin, Dadashi, and Hussenot}]{gemmateam2025gemma3technicalreport}
Gemma Team, Aishwarya Kamath, Johan Ferret, Shreya Pathak, Nino Vieillard, Ramona Merhej, Sarah Perrin, Tatiana Matejovicova, Alexandre Ramé, Morgane Rivière, Louis Rouillard, Thomas Mesnard, Geoffrey Cideron, Jean bastien Grill, Sabela Ramos, Edouard Yvinec, Michelle Casbon, Etienne Pot, Ivo Penchev, Gaël Liu, Francesco Visin, Kathleen Kenealy, Lucas Beyer, Xiaohai Zhai, Anton Tsitsulin, Robert Busa-Fekete, Alex Feng, Noveen Sachdeva, Benjamin Coleman, Yi~Gao, Basil Mustafa, Iain Barr, Emilio Parisotto, David Tian, Matan Eyal, Colin Cherry, Jan-Thorsten Peter, Danila Sinopalnikov, Surya Bhupatiraju, Rishabh Agarwal, Mehran Kazemi, Dan Malkin, Ravin Kumar, David Vilar, Idan Brusilovsky, Jiaming Luo, Andreas Steiner, Abe Friesen, Abhanshu Sharma, Abheesht Sharma, Adi~Mayrav Gilady, Adrian Goedeckemeyer, Alaa Saade, Alex Feng, Alexander Kolesnikov, Alexei Bendebury, Alvin Abdagic, Amit Vadi, András György, André~Susano Pinto, Anil Das, Ankur Bapna, Antoine Miech, Antoine Yang, Antonia Paterson, Ashish
  Shenoy, Ayan Chakrabarti, Bilal Piot, Bo~Wu, Bobak Shahriari, Bryce Petrini, Charlie Chen, Charline~Le Lan, Christopher~A. Choquette-Choo, CJ~Carey, Cormac Brick, Daniel Deutsch, Danielle Eisenbud, Dee Cattle, Derek Cheng, Dimitris Paparas, Divyashree~Shivakumar Sreepathihalli, Doug Reid, Dustin Tran, Dustin Zelle, Eric Noland, Erwin Huizenga, Eugene Kharitonov, Frederick Liu, Gagik Amirkhanyan, Glenn Cameron, Hadi Hashemi, Hanna Klimczak-Plucińska, Harman Singh, Harsh Mehta, Harshal~Tushar Lehri, Hussein Hazimeh, Ian Ballantyne, Idan Szpektor, Ivan Nardini, Jean Pouget-Abadie, Jetha Chan, Joe Stanton, John Wieting, Jonathan Lai, Jordi Orbay, Joseph Fernandez, Josh Newlan, Ju~yeong Ji, Jyotinder Singh, Kat Black, Kathy Yu, Kevin Hui, Kiran Vodrahalli, Klaus Greff, Linhai Qiu, Marcella Valentine, Marina Coelho, Marvin Ritter, Matt Hoffman, Matthew Watson, Mayank Chaturvedi, Michael Moynihan, Min Ma, Nabila Babar, Natasha Noy, Nathan Byrd, Nick Roy, Nikola Momchev, Nilay Chauhan, Noveen Sachdeva, Oskar
  Bunyan, Pankil Botarda, Paul Caron, Paul~Kishan Rubenstein, Phil Culliton, Philipp Schmid, Pier~Giuseppe Sessa, Pingmei Xu, Piotr Stanczyk, Pouya Tafti, Rakesh Shivanna, Renjie Wu, Renke Pan, Reza Rokni, Rob Willoughby, Rohith Vallu, Ryan Mullins, Sammy Jerome, Sara Smoot, Sertan Girgin, Shariq Iqbal, Shashir Reddy, Shruti Sheth, Siim Põder, Sijal Bhatnagar, Sindhu~Raghuram Panyam, Sivan Eiger, Susan Zhang, Tianqi Liu, Trevor Yacovone, Tyler Liechty, Uday Kalra, Utku Evci, Vedant Misra, Vincent Roseberry, Vlad Feinberg, Vlad Kolesnikov, Woohyun Han, Woosuk Kwon, Xi~Chen, Yinlam Chow, Yuvein Zhu, Zichuan Wei, Zoltan Egyed, Victor Cotruta, Minh Giang, Phoebe Kirk, Anand Rao, Kat Black, Nabila Babar, Jessica Lo, Erica Moreira, Luiz~Gustavo Martins, Omar Sanseviero, Lucas Gonzalez, Zach Gleicher, Tris Warkentin, Vahab Mirrokni, Evan Senter, Eli Collins, Joelle Barral, Zoubin Ghahramani, Raia Hadsell, Yossi Matias, D.~Sculley, Slav Petrov, Noah Fiedel, Noam Shazeer, Oriol Vinyals, Jeff Dean, Demis Hassabis,
  Koray Kavukcuoglu, Clement Farabet, Elena Buchatskaya, Jean-Baptiste Alayrac, Rohan Anil, Dmitry, Lepikhin, Sebastian Borgeaud, Olivier Bachem, Armand Joulin, Alek Andreev, Cassidy Hardin, Robert Dadashi, and Léonard Hussenot. 2025.
\newblock \href {https://arxiv.org/abs/2503.19786} {Gemma 3 technical report}.
\newblock \emph{Preprint}, arXiv:2503.19786.

\bibitem[{Touvron et~al.(2023)Touvron, Lavril, Izacard, Martinet, Lachaux, Lacroix, Rozière, Goyal, Hambro, Azhar, Rodriguez, Joulin, Grave, and Lample}]{Touvron2023}
Hugo Touvron, Thibaut Lavril, Gautier Izacard, Xavier Martinet, Marie-Anne Lachaux, Timothée Lacroix, Baptiste Rozière, Naman Goyal, Eric Hambro, Faisal Azhar, Aurelien Rodriguez, Armand Joulin, Edouard Grave, and Guillaume Lample. 2023.
\newblock \href {https://arxiv.org/abs/2302.13971v1} {Llama: Open and efficient foundation language models}.

\bibitem[{Wallach et~al.(2025)Wallach, Desai, Cooper, Wang, Atalla, Barocas, Blodgett, Chouldechova, Corvi, Dow et~al.}]{wallach2025position}
Hanna Wallach, Meera Desai, A~Feder Cooper, Angelina Wang, Chad Atalla, Solon Barocas, Su~Lin Blodgett, Alexandra Chouldechova, Emily Corvi, P~Alex Dow, et~al. 2025.
\newblock Position: Evaluating generative ai systems is a social science measurement challenge.
\newblock \emph{arXiv preprint arXiv:2502.00561}.

\bibitem[{Wang et~al.(2024)Wang, Cheng, Zhan, Li, Song, and Liu}]{wang2024openchatadvancingopensourcelanguage}
Guan Wang, Sijie Cheng, Xianyuan Zhan, Xiangang Li, Sen Song, and Yang Liu. 2024.
\newblock \href {https://arxiv.org/abs/2309.11235} {Openchat: Advancing open-source language models with mixed-quality data}.
\newblock \emph{Preprint}, arXiv:2309.11235.

\bibitem[{Wankhade et~al.(2022)Wankhade, Rao, and Kulkarni}]{Wankhade2022}
Mayur Wankhade, Annavarapu Chandra~Sekhara Rao, and Chaitanya Kulkarni. 2022.
\newblock \href {https://doi.org/10.1007/S10462-022-10144-1} {A survey on sentiment analysis methods, applications, and challenges}.
\newblock \emph{Artificial Intelligence Review 2022 55:7}, 55:5731--5780.

\bibitem[{Xu et~al.(2025)Xu, Lawrence, Dubey, Pandey, Falck, Ueno, Nori, Sharma, Sharma, and González}]{xu2025re-imagine}
Xinnuo Xu, Rachel Lawrence, Kshitij Dubey, Atharva Pandey, Fabian Falck, Risa Ueno, Aditya Nori, Rahul Sharma, Amit Sharma, and Javier González. 2025.
\newblock \href {https://www.microsoft.com/en-us/research/publication/re-imagine-symbolic-benchmark-synthesis-for-reasoning-evaluation/} {Re-imagine: Symbolic benchmark synthesis for reasoning evaluation}.
\newblock In \emph{ICLR 2025 - Workshop on Reasoning and Planning for LLMs}.

\bibitem[{Yang et~al.(2021)Yang, Li, Cunningham, Zhang, Smyth, and Dong}]{yang-etal-2021-exploring}
Linyi Yang, Jiazheng Li, Padraig Cunningham, Yue Zhang, Barry Smyth, and Ruihai Dong. 2021.
\newblock \href {https://doi.org/10.18653/v1/2021.acl-long.26} {Exploring the efficacy of automatically generated counterfactuals for sentiment analysis}.
\newblock In \emph{Proceedings of the 59th Annual Meeting of the Association for Computational Linguistics and the 11th International Joint Conference on Natural Language Processing (Volume 1: Long Papers)}, pages 306--316, Online. Association for Computational Linguistics.

\bibitem[{Yoo and Rayz(2021)}]{yoo2021understanding}
Byungkyu Yoo and Julia~Taylor Rayz. 2021.
\newblock Understanding emojis for sentiment analysis.
\newblock In \emph{The international FLAIRS conference proceedings}, volume~34.

\bibitem[{Zhang(2024)}]{Zhang2024sentiment}
Junfeng Zhang. 2024.
\newblock Sentiment and language: A socio-semiotic analysis.
\newblock \emph{Philosophy Journal}, 3(1):118--127.

\bibitem[{Zhang et~al.(2023)Zhang, Cahyawijaya, Cruz, Winata, and Aji}]{zhang-etal-2023-multilingual}
Ruochen Zhang, Samuel Cahyawijaya, Jan Christian~Blaise Cruz, Genta Winata, and Alham Aji. 2023.
\newblock \href {https://doi.org/10.18653/v1/2023.emnlp-main.774} {Multilingual large language models are not (yet) code-switchers}.
\newblock In \emph{Proceedings of the 2023 Conference on Empirical Methods in Natural Language Processing}, pages 12567--12582, Singapore. Association for Computational Linguistics.

\end{thebibliography}

\clearpage
\appendix
\section{Appendix}
\subsection{Annotation Protocol}
\label{appendix:annotation}

\textbf{Annotators and Process.}  
This protocol was developed to guide consistent sentiment annotation of informal, multilingual WhatsApp messages exchanged among youth in Nairobi. Two trained annotators—both fluent in English, Swahili, and Sheng—applied the protocol over the course of one month. Annotation covered 6,197 messages drawn from a dataset of informal, code-mixed conversations among youth living with HIV. Work was conducted in Excel. Annotators labeled each message independently, treating it as a standalone utterance while considering cultural context, code-switching, and emoji use. We began with a jointly labeled calibration set of 100 examples, followed by independent annotation with regular meetings to discuss edge cases and resolve ambiguities.

\textbf{Sentiment Categories.}  
Messages were labeled as \textbf{Negative (-1)}, \textbf{Neutral (0)}, or \textbf{Positive (1)} based on expressed affect. Annotators were instructed to:

\begin{itemize}
  \item Label as \textbf{Negative} if the message expressed frustration, sadness, criticism, or distress (e.g., \textit{"I am tired !!!", "You have a mental problem"}).
  \item Label as \textbf{Neutral} if the message conveyed information, routine conversation, or general greetings without strong sentiment (e.g., \textit{"When are you coming?", "Good morning ~\raisebox{-0.2em}{\includegraphics[height=1em]{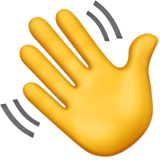}}"}).
  \item Label as \textbf{Positive} if the message expressed joy, support, pride, or optimism (e.g., \textit{"I'm much happy to interact and share with you guys!"}).
\end{itemize}

\textbf{Ambiguity and Cultural Nuance.}  
Annotators flagged ambiguous cases with written justifications. Given the culturally grounded and multilingual nature of the data, particular attention was paid to tone, idioms, emoji use, and context-specific expressions of affect.

%%%%%%%%DATASTATISTICS
\subsection{Evaluation Subsets}

\begin{table}[H]
\centering
\scriptsize
\begin{tabular}{|m{1.95cm}|m{0.85cm}|m{0.97cm}|m{0.85cm}|c|}
\toprule
\textbf{Subset} & \textbf{Positive} & \textbf{Negative} & \textbf{Neutral} & \textbf{Total} \\
\midrule
Gold Set & 1196 & 351 & 4574 & 6,121 \\
\midrule
Synthetic Set & 351 & 1196 & - & 1,547 \\
\midrule
Ambiguous Set & - & - & - & 76 \\
\bottomrule
\end{tabular}
\caption{Sentiment-wise distribution of messages.}
\label{tab:sentiment-subset-breakdown}
\end{table}

%%%%%LLMS PERFORMANCE

\subsection{Overall Model Performance}
\begin{table}[H]
\centering
\scriptsize
\setlength{\tabcolsep}{6pt}
\begin{tabular}{lccccc}
\toprule
\textbf{Model} & \textbf{Pos} & \textbf{Neg} & \textbf{Neu} & \textbf{Avg} & \textbf{Cov. \%} \\
\midrule
\multicolumn{6}{l}{\textbf{\textit{Gold Set (annotated Pos/Neg/Neu)}}} \\
\texttt{GPT-4-Turbo}     & 0.98 & 0.92 & 0.75 & 0.88 & 100.0 \\
\texttt{GPT-4-32k}       & 0.93 & 0.90 & 0.86 & 0.90 & 100.0 \\
\texttt{Gemma-3-27B}     & 0.93 & 0.96 & 0.79 & 0.89 & 100.0 \\
\texttt{Phi-4}           & 0.93 & 0.91 & 0.80 & 0.88 & 100.0 \\
\texttt{Mistral-7B}      & 0.91 & 0.88 & 0.92 & 0.90 & 98.9  \\
\texttt{OpenChat-3.5}    & 0.93 & 0.77 & 0.87 & 0.86 & 99.9  \\
\texttt{LLaMA-3-8B}      & 0.94 & 0.51 & 0.86 & 0.77 & 92.9  \\
\midrule
\multicolumn{6}{l}{\textbf{\textit{Pre-CF (original Pos/Neg examples)}}} \\
\texttt{GPT-4-Turbo}     & 0.98 & 0.94 & —    & 0.96 & 100.0 \\
\texttt{GPT-4-32k}       & 0.99 & 0.95 & —    & 0.97 & 100.0 \\
\texttt{Gemma-3-27B}     & 0.97 & 0.90 & —    & 0.94 & 100.0 \\
\texttt{Phi-4}           & 0.97 & 0.90 & —    & 0.94 & 100.0 \\
\texttt{OpenChat-3.5}    & 0.97 & 0.85 & —    & 0.91 & 100.0 \\
\texttt{Mistral-7B}      & 0.97 & 0.89 & —    & 0.93 & 95.9  \\
\texttt{LLaMA-3-8B}      & 0.97 & 0.77 & —    & 0.87 & 90.2  \\
\midrule
\multicolumn{6}{l}{\textbf{\textit{Post-CF (synthetic counterfactuals)}}} \\
\texttt{GPT-4-Turbo}     & 0.97 & 0.99 & —    & 0.98 & 100.0 \\
\texttt{GPT-4-32k}       & 0.97 & 0.99 & —    & 0.98 & 100.0 \\
\texttt{Phi-4}           & 0.67 & 0.90 & —    & 0.79 & 99.5  \\
\texttt{Gemma-3-27B}     & 0.97 & 0.99 & —    & 0.98 & 47.6  \\
\texttt{Mistral-7B}      & 0.97 & 0.99 & —    & 0.98 & 47.6  \\
\texttt{OpenChat-3.5}    & 0.90 & 0.97 & —    & 0.93 & 47.4  \\
\texttt{LLaMA-3-8B}      & 0.91 & 0.96 & —    & 0.93 & 37.6  \\
\bottomrule
\end{tabular}
\caption{F1 scores by sentiment class on the Gold Set, Pre-CF (original positive/negative examples used to generate counterfactuals), and Post-CF (synthetic counterfactuals with flipped sentiment). Coverage rate (\%) reflects the proportion of examples for which a model returned a valid sentiment label.}
\label{tab:quant-performance}
\end{table}

%%%%%%
\subsection{Prediction Agreement Across Models}
\begin{figure}[H]
    \centering
    \includegraphics[width=1\linewidth]{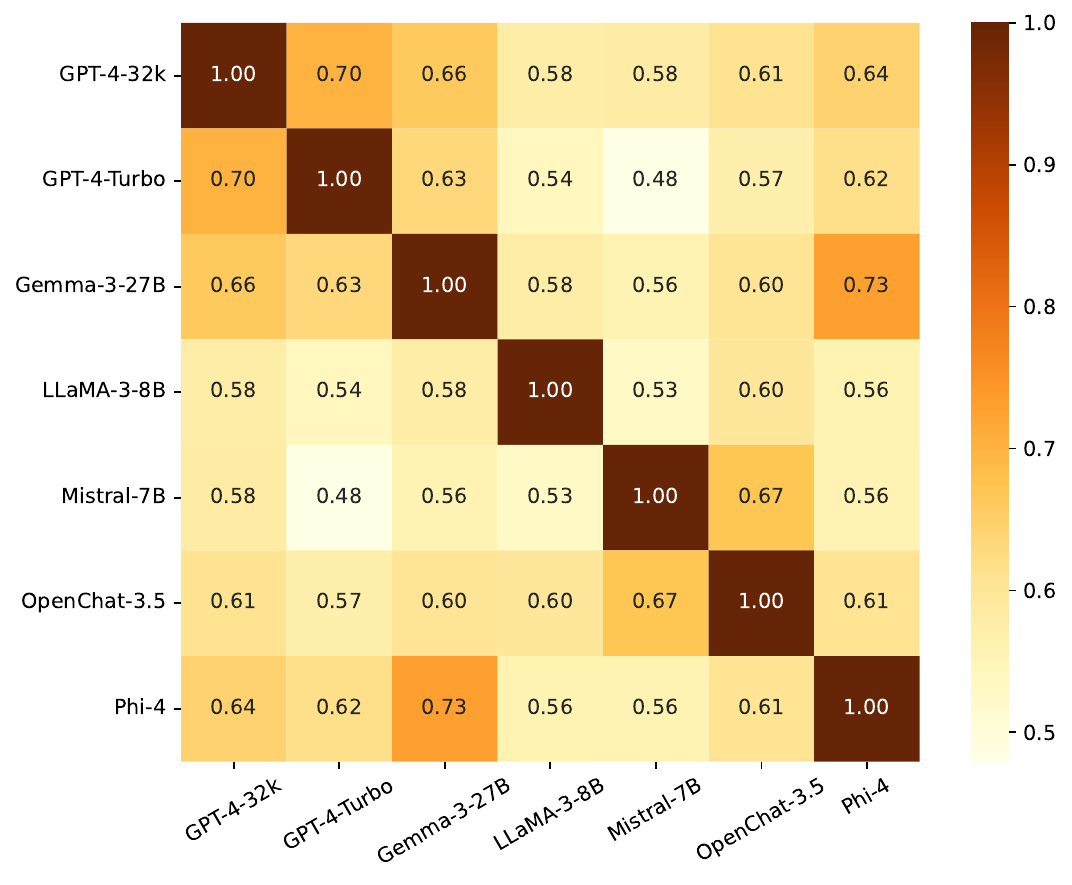}
    \caption{Cohen’s $\kappa$ agreement between model predictions on the Gold Set. Despite full annotator agreement, models show only moderate pairwise consistency—indicating divergence in their underlying reasoning and sensitivity to sentiment cues}
    \label{fig:model-agreement}
\end{figure}

%%%%%%%%PROMPTS%%%%%%%%
\subsection{Model Prompts}

\begin{table}[H]
\centering
\scalebox{0.72}{
\begin{tabular}{p{110mm}}
\toprule
\textbf{Sentiment Classification + Explanation Prompt} \\
\texttt{You are an NLP assistant for sentiment analysis.} \\
\texttt{Given a WhatsApp message (QUERY), classify its sentiment as Positive, Negative, or Neutral.} \\
\texttt{Provide a justification using extracted keywords and a brief explanation.} \\
\texttt{Return your confidence score (0–5). Use JSON output format only.} \\

\\
\texttt{The prompt includes:} \\
\texttt{- Sentiment definitions (Positive, Neutral, Negative)} \\
\texttt{- Examples of clearly and ambiguously labeled messages} \\
\texttt{- JSON output format with keywords, explanation, label, and confidence} \\

\\
\texttt{QUERY: "\{query\}"} \\

\\
\texttt{Output Format:} \\
\texttt{\{} \\
\texttt{~~"justification": \{} \\
\texttt{~~~~"keywords": [ ... ],} \\
\texttt{~~~~"explanation": "..." \},} \\
\texttt{~~"sentiment": "...",} \\
\texttt{~~"confidence\_score": "..." } \\
\texttt{\}} \\

\\
\bottomrule
\end{tabular}
}
\caption{Instruction prompt for joint sentiment classification, justification, and confidence scoring.}
\label{tab:sentiment_prompt}
\end{table}
%%%%%%%%%Counterfactual Quality Evaluation Prompt (LLM-as-Judge)
\begin{table}[H]
\centering
\scalebox{0.75}{
\begin{tabular}{p{100mm}}
\toprule
\textbf{Counterfactual Evaluation Prompt} \
\texttt{You are evaluating a synthetic (GPT-4-generated) version of a WhatsApp message.} \
\texttt{The synthetic message is a sentiment-flipped version of the original.} \

\texttt{Assess the quality of the synthetic message along four criteria using 0 or 1:} \\ \texttt{1. Fluency — Is the synthetic message grammatically correct and readable?} \\ \texttt{2. Naturalness — Does it sound plausible for a human to write?} \\ \texttt{3. Sentiment Flip Clarity — Is the sentiment clearly flipped from the original?} \\ \texttt{4. Meaning Preservation — Is the core meaning preserved aside from the sentiment?} \\ \\ \texttt{Original Message: "\{original\}"} \\ \texttt{Synthetic Message: "\{flipped\}"} \\ \texttt{Transformation Type: \{transformation\}} \\ \texttt{GPT-4 Explanation for the Flip: "\{explanation\}"} \\ \\ \texttt{Return ONLY this JSON:} \\ \texttt{\{} \\ \texttt{~~"fluency": 0 or 1,} \\ \texttt{~~"naturalness": 0 or 1,} \\ \texttt{~~"sentiment\_flip\_clarity": 0 or 1,} \\ \texttt{~~"meaning\_preservation": 0 or 1,} \\ \texttt{~~"annotator\_comment": "optional comment (string)"} \\ \texttt{\}} \\ \bottomrule \end{tabular} } \caption{Prompt used to evaluate quality of synthetic counterfactuals across four rubric dimensions.} 
\label{tab:cf_quality_prompt} 
\end{table}

%%%%%CombinedCF generator
\begin{table}[htbp]
\centering
\scalebox{0.65}{
\begin{tabular}{p{120mm}}
\toprule
\textbf{(a) Counterfactual Generation Prompt} \\
\texttt{You are an NLP assistant helping researchers generate high-quality counterfactual examples for sentiment classification.} \\
\texttt{Given a WhatsApp-style message and its sentiment (Positive or Negative), generate 3 distinct versions that flip the sentiment.} \\
\texttt{Only modify necessary components. Preserve fluency and realism. Respect informal tone.} \\

\texttt{You may flip sentiment by changing components such as:} \\
\texttt{- keywords, phrases, negation, intent framing, tone (e.g., sarcasm), sentiment valence, emojis/icons, code-mixing} \\

\texttt{Input:} \\
\texttt{Original message: "\{original\_message\}"} \\
\texttt{Original sentiment: "\{original\_sentiment\}"} \\

\texttt{Output Format (JSON List of 3 Objects):} \\
\texttt{\{} \\
\texttt{~~"cf\_text": "...",} \\
\texttt{~~"components\_changed": [...],} \\
\texttt{~~"flip\_explanation": "..."} \\
\texttt{\}} \\

\\
\midrule
\textbf{(b) Counterfactual Filtering Prompt} \\
\texttt{You are a sentiment evaluation assistant. Your task is to select the best counterfactual rewrite of a message.} \\

\texttt{ORIGINAL MESSAGE} \\
\texttt{"\{original\}"} \\
\texttt{(Sentiment: \{original\_sentiment\})} \\

\texttt{COUNTERFACTUAL CANDIDATES} \\
\texttt{1. "\{cf1\}"} \\
\texttt{2. "\{cf2\}"} \\
\texttt{3. "\{cf3\}"} \\

\texttt{INSTRUCTIONS} \\
\texttt{Your goal is to identify which counterfactual most effectively flips the sentiment while remaining realistic and fluent.} \\
\texttt{- Flip sentiment plausibly} \\
\texttt{- Sound natural in WhatsApp chat} \\
\texttt{- Preserve meaning/context where possible} \\

\texttt{RESPONSE FORMAT (JSON only):} \\
\texttt{\{} \\
\texttt{~~"selected\_cf": "...",} \\
\texttt{~~"justification": "...",} \\
\texttt{~~"predicted\_sentiment": "Positive / Negative"} \\
\texttt{\}} \\

\bottomrule
\end{tabular}
}
\caption{Combined prompts for generating and selecting counterfactual sentiment flips.}
\label{tab:cf_prompts}
\end{table}

%%%%%%%Explanation Evaluation Prompt (LLM-as-Judge)
\begin{table}[htbp]
% \begin{table}[H]
\centering
\scalebox{0.75}{
\begin{tabular}{p{100mm}}
\toprule
\textbf{Explanation Evaluation Prompt} \
\texttt{You are a language model tasked with evaluating the quality of a sentiment explanation.} \
\texttt{Evaluate the explanation for the following:} \
\texttt{1. Faithfulness – Does it reflect the original message and prediction without hallucinating?} \
\texttt{2. Contextual Appropriateness – Is it culturally and linguistically aware?} \
\texttt{3. Logical Coherence – Is it internally consistent and justified?} \
\texttt{4. Clarity and Completeness – Is it clear, specific, and sufficient?} \

\texttt{Message:} \\ \texttt{"\{message\}"} \\ \texttt{Predicted Sentiment: \{prediction\}} \\ \texttt{Explanation: "\{explanation\}"} \\ \\ \texttt{Return ONLY this JSON:} \\ \texttt{\{} \\ \texttt{~~"faithfulness": 0 or 1,} \\ \texttt{~~"contextual\_appropriateness": 0 or 1,} \\ \texttt{~~"logical\_coherence": 0 or 1,} \\ \texttt{~~"clarity\_and\_completeness": 0 or 1,} \\ \texttt{~~"annotator\_comment": "optional comment (string)"} \\ \texttt{\}} \\ \bottomrule \end{tabular} } 
\caption{Prompt used to evaluate explanation quality across four rubric dimensions.} 
\label{tab:explanation_eval_prompt} 
\end{table}

\onecolumn
% \section{Appendix}

%%%%%% CULTURAL NUANCE EXAMPLES
\subsection{Further Examples from Our WhatsApp Dataset: Cultural Nuance and Annotator Disagreement}
\label{appendix:cultural-nuance}

\begin{table}[H]
    \centering
    \small
    \begin{tabular}{|m{5cm}|m{10cm}|}
    \toprule
    \textbf{Example} & \textbf{Explanation} \\
    \midrule

    \textit{My friends it was heard to take drugs bt i just take heart} &
    \begin{itemize}[leftmargin=*, itemsep=1pt, topsep=0pt]
        \item Can be read differently due to \textit{situational context (sympathy)}.
        \item Shows \textbf{emotional vulnerability}, which may invite empathy or humor depending on setting.
        \item Use of \textbf{``take heart''} is culturally influenced—often heard in African English as a way to express resilience.
        \item The spelling (``heard'' instead of ``hard'') could be \textbf{interpreted differently} (innocent typo vs. deeper linguistic variation).
    \end{itemize} \\
    \midrule

    \textit{Kama hauko School shindaapo} \newline “Even you are not in school just stay there” &
    \begin{itemize}[leftmargin=*, itemsep=1pt, topsep=0pt]
        \item Can be read differently due to \textit{schooling context}.
        \item Often used \textbf{sarcastically or dismissively}, especially in online chat.
        \item The phrase can also reflect \textbf{class-based or knowledge-based exclusion} (``If you're not educated, stay out of this'').
        \item Code-mixing adds a \textbf{layer of urban youth culture} and localized meaning.
    \end{itemize} \\
    \midrule

    \textit{He is faithful all the time} &
    \begin{itemize}[leftmargin=*, itemsep=1pt, topsep=0pt]
        \item Can be read differently due to \textit{religion}.
        \item Common in \textbf{Christian communities}, especially in African contexts—often part of a \textbf{call-and-response}.
        \item Can express \textbf{faith during suffering}, giving it emotional depth in testimonies or public speeches.
        \item Without context, it may be misread as a general statement about a person rather than a \textbf{declaration about God}.
    \end{itemize} \\
    \bottomrule
    \end{tabular}
    \caption{Examples of cultural nuance and their context-dependent interpretations.}
    \label{tab:cultural-nuance}
\end{table}

%%%%%%RUBRICS
\subsection{Rubrics for Evaluation}
\label{appendix:rubrics}
\subsubsection{Explanation Evaluation Rubric}

Each model-generated explanation was evaluated along four binary (0/1) dimensions:

\begin{itemize}
    \item \textbf{Faithfulness:} Does the explanation accurately reflect the input message and how it informed the model’s sentiment prediction? Explanations that include hallucinated, fabricated, or unrelated content should be scored \textbf{0}.
    \item \textbf{Contextual Appropriateness:} Does the explanation show awareness of cultural, social, or linguistic context? If it fails to address relevant tone, code-mixing, or local expressions, assign \textbf{0}. Optional comments may highlight cultural or linguistic mismatches.
    \item \textbf{Logical Coherence:} Is the explanation internally consistent and logically connected to the sentiment label? Contradictory or illogical justifications are scored \textbf{0}.
    \item \textbf{Clarity and Completeness:} Is the explanation clear, specific, and sufficient to support the sentiment label? Vague or underspecified rationales receive \textbf{0}.
\end{itemize}

\noindent\textit{Scoring:} 1 = Yes; 0 = No \\
\textit{Note:} Annotators were asked to leave optional comments when assigning a score of 0, especially for cultural/contextual errors or hallucinations.

\vspace{1em}
\subsubsection{Synthetic Data Evaluation Rubric}

Each GPT-4-generated counterfactual message was evaluated using the following binary (0/1) criteria:

\begin{itemize}
    \item \textbf{Fluency:} Is the synthetic message grammatically well-formed and fluent?
    \item \textbf{Naturalness:} Does the message sound plausible or likely to have been written by a real user?
    \item \textbf{Sentiment Flip Clarity:} Is the reversal in sentiment (compared to the original message) clear and consistent?
    \item \textbf{Meaning Preservation:} Aside from sentiment, does the core meaning/topic of the original message remain intact? Large semantic shifts receive \textbf{0}.
\end{itemize}

\noindent\textit{Scoring:} 1 = Yes; 0 = No \\
\textit{Note:} Annotators were encouraged to flag particularly good or bad examples, especially where tone, fluency, or cultural grounding were notably off.

% \newpage
%%%%%%%%TAXONOMY
\subsection{Sentiment Transformation Taxonomy}
\label{appendix:transformations}

To guide counterfactual generation, we organize sentiment-altering edits into the following transformation types:

\begin{table}[h]
\centering
\scriptsize
% \small
\begin{tabular}{p{3.2cm} p{4.2cm} p{5.5cm}}
\toprule
\textbf{Transformation Type} & \textbf{Definition} & \textbf{Example} \\
\midrule
Negation & Add or remove negation to reverse sentiment. & \textit{“I like it”} $\rightarrow$ \textit{“I don't like it”} \\
\addlinespace
Tone / Intent Shift & Change the tone or implied intent of the message. & \textit{“You could do better”} $\rightarrow$ \textit{“You're doing great”} \\
\addlinespace
Emoji Substitution & Replace emoji's to reflect different sentiment.
& \textit{“\raisebox{-0.2em}{\includegraphics[height=1em]{images/sweat_smile.png}}”} 
$\rightarrow$ 
\textit{“\raisebox{-0.2em}{\includegraphics[height=1em]{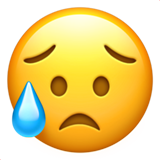}}”} \\
\addlinespace
Keyword Substitution & Swap a sentiment-bearing word. & \textit{“Useful advice”} $\rightarrow$ \textit{“Terrible advice”} \\
\addlinespace
Phrase Rewording & Paraphrase to shift sentiment while preserving meaning. & \textit{“You always help me”} $\rightarrow$ \textit{“You always get in my way”} \\
\bottomrule
\end{tabular}
\caption{Taxonomy of sentiment-altering transformations used in counterfactual generation.}
\label{tab:transformation-taxonomy}
\end{table}

\begin{figure} [H]
    \centering
    \includegraphics[width=0.75\linewidth]{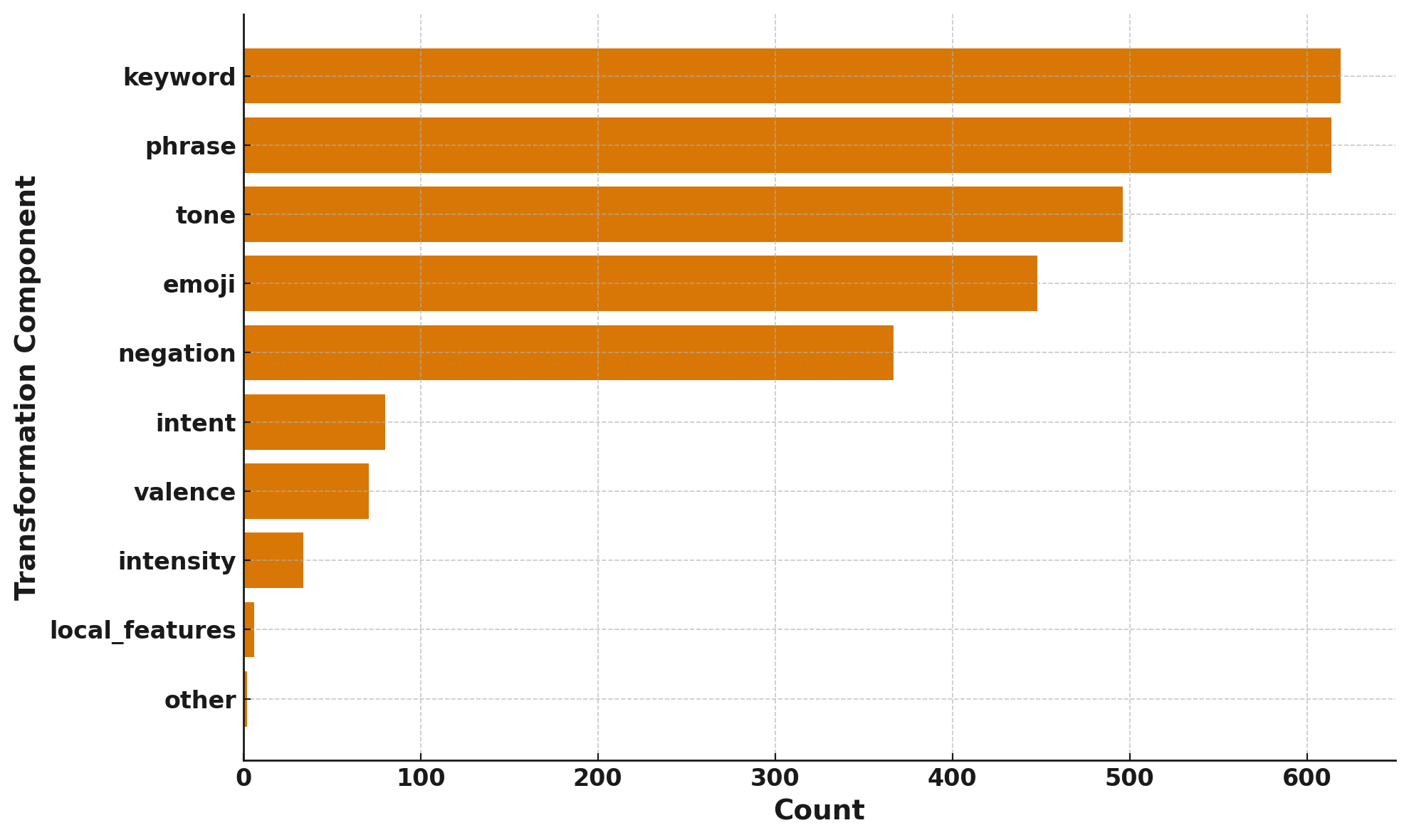}
    \caption{Distribution of transformation components used to generate counterfactual examples. Most sentiment flips involved keyword or phrase substitutions, followed by tone shifts, emoji manipulation, and negation. Less frequent transformations include intent, valence, and intensity adjustments.}
    \label{fig:transformations}
\end{figure}
%%%%%%%%

%%%%%%%%SYNTHETIC DATA EXAMPLES

\subsection{Outputs of Synthetic Experiments}

%%%%CONCISTENCY
\subsubsection{Component Consistency}
\label{appendix:synthetic-consistency}

\begin{table}[H]
\centering
\small
\begin{tabular}{|m{3cm}|m{3cm}|m{2.5cm}|m{6cm}|}
\toprule
\textbf{Original message} & \textbf{Flipped message} & \textbf{LLM identified components} & \textbf{LLM rationale} \\
\midrule

Hahaha (+) & This is not funny at all. (–) & phrase, \textcolor{red}{valence} & Changing ‘emoji why?’ to a direct statement expressing displeasure flips the sentiment from positive to negative. \textcolor{red}{Note: hallucinated explanantion}\\
\midrule

Hahaha (+) & not funny at all (–) & \textcolor{red}{keyword}, negation & Changing 'hahaha' to 'not funny at all' adds negation and alters the \textcolor{red}{key phrase} to express dissatisfaction, flipping the sentiment to negative. \\
\midrule

Ahahhaa (+) & Not funny at all. (–) & \textcolor{red}{keyword}, negation & Adding negation and changing the keyword from a positive laughter expression to a negative critique directly flips the sentiment. \\
\midrule

Haha (+) & not funny (–) & phrase, negation & Introducing ‘not’ before ‘funny’ adds a negation to the original message, shifting the sentiment from amusement to disapproval. \\
\midrule

Hahaaaaa addemoji & Ughhh addemoji & emoji, tone & Changing the emoji from addemoji (happy) to addemoji (annoyed) and the exaggerated sound from ‘Hahaaaaa’ to ‘Ughhh’ shifts the sentiment from positive to negative. \\
\midrule

hahaha addemoji & this is not funny at all addemoji & phrase, emoji \textcolor{red}{[no mentioning of negation]} & Adding text that explicitly states discontent (‘this is not funny at all’) and using an angry emoji changes the sentiment from amusement to displeasure. \\
\bottomrule
\end{tabular}
\caption{Examples from our synthetic experiment showing varied LLM component sensitivity and consistency in handling laughter expressions.}
\label{tab:synthetic-consistency}
\end{table}

%%%%%%WORKEDWELL
\subsubsection{What Worked Well}
\label{appendix:synthetic-success}

\begin{table}[H]
\centering
\small
\begin{tabular}{|m{3.5cm}|m{3.5cm}|m{2.5cm}|m{4.5cm}|}
\toprule
\textbf{Original message} & \textbf{Flipped message} & \textbf{LLM identified components} & \textbf{LLM rationale} \\
\midrule

\textit{Pia mi nko poa} (+) \newline ``Me too, I’m okay.'' & \textit{Pia mi siko poa kabisa leo} (–) \newline ``Me too, I’m really not okay today.'' & phrase, intensifier & Adding `siko' (not) and `kabisa' (at all) to the original message intensifies the negative sentiment, flipping it from positive. \newline [\textbf{Reflection}: Correct flip with Standard Swahili] \\
\midrule

\textit{Ata me wananiboo leo yote hawajatext} (–) \newline Even me, they’re boring me the whole day they haven’t texted. & \textit{Ata me leo nimefurahishwa, wamekuwa wakinitext mara kwa mara.} (–) \newline ``Even me, today I’ve been made happy, they’ve been texting me again and again.'' & keywords, phrase & By modifying the phrase to express happiness and frequent communication, the sentiment changes from negative to positive. \newline [\textbf{Reflection}: Correct flip with Standard Swahili]\\
\midrule

\textit{I will check on that coz nakohoa sana adi naumwa na kifua} (–) \newline 
``I’ll check on that because I’m coughing a lot - to the point my chest hurts.''
 &

I will check on that coz I’m feeling much better now, the cough and chest pain are gone! (+) & keywords, valence & The switch from describing ongoing symptoms to stating that the symptoms are gone (‘the cough and chest pain are gone’) changes the sentiment from negative to positive by indicating an improvement in health. \newline [\textbf{Reflection}: Correct flip with English Translation]\\
\bottomrule
\end{tabular}
\caption{Examples from our synthetic experiment showing successful sentiment flips and clear rationale.}
\label{tab:synthetic-success}
\end{table}

%%%%%WORKEDLESS
\subsubsection{What Worked Less Well}
\label{appendix:synthetic-failures}

\begin{table}[H]
\centering
\small
\begin{tabular}{|p{3cm}|p{3.8cm}|p{4.5cm}|p{3.3cm}|}
\toprule
 & \textbf{Example 1} & \textbf{Example 2} & \textbf{Example 3} \\
\midrule

\textbf{Original message} & 
\textit{but kulinice} \newline ``But it was nice'' & 
And thank you 5017 for very \textbf{clear} elaboration about the questions at hand. \textbf{I love} how you have handled and answered them \textbf{to the point} (+) & 
Dia hope you will be coming visiting I enjoyed your company alot your fun to be with (+) \\
\midrule

\textbf{Flipped message} & 
but \textit{kulinice} was disappointing & 
And thanks 5017 for very \textbf{vague} explanations about the questions at hand. \textbf{I hate} how you have \textbf{mishandled and dodged them completely}. (–) & 
Dia, please don't come over. Honestly, hanging out was more exhausting than enjoyable, you're not that fun to be with. (–) \\
\midrule

\textbf{LLM identified \newline components} & 
key phrase & 
keywords, key phrases, tone & 
negation, tone, key phrases \\
\midrule

\textbf{LLM rationale} & 
Changing 'kulinice' to 'kulinice was disappointing' directly flips the sentiment by expressing a negative experience. & 
Changing 'clear elaboration' to 'vague explanations' and 'love' to 'hate' directly flips the positive sentiment to negative. Additionally, 'handled and answered them to the point' was changed to 'mishandled and dodged them completely' to emphasize dissatisfaction. & 
Introducing a request for absence ('please don't come over') and directly stating negative feelings ('more exhausting than enjoyable', 'not that fun') shifts the sentiment from positive to negative. \\
\midrule

\textbf{Reflections / Insights} & 
From the LLM’s explanation sw-en codemixed \textit{kulinice} meaning ``it was nice'' misunderstood; term ``kulinice'' was ignored rather than changed because it could not handle the code-mix expression. & 
Flipped message feels somewhat unnatural: the ``thanks'' gets in the way of creating weird sarcasm/strong negativity with words like ``hate''; hate may be the flip to ``love'' but it’s rare to have such expressions in feedback about questions. & 
The flipped message sounds strong/extreme; unnatural. \\
\bottomrule
\end{tabular}
\caption{Examples from our synthetic experiment showing less effective sentiment flips.}
\label{tab:synthetic-failures}
\end{table}
%%%%%%%%

%%%%%%%%EXP QUALITY
\subsection{Model Explanation Quality, Confidence, and Coverage}
\begin{table}[H]
\centering
\scriptsize
\begin{tabular}{llcccc}
\toprule
\textbf{Dataset} & \textbf{Model} & \textbf{Faithfulness} & \textbf{Contextual Approp.} & \textbf{Logical Coherence} & \textbf{Clarity \& Completeness} \\
\midrule
\multirow{3}{*}{Ambiguous Set} 
 & GPT-4-32k        & 1.000 & 0.950 & 1.000 & 0.950 \\
 & GPT-4-Turbo      & 1.000 & 1.000 & 1.000 & 1.000 \\
 & Gemma-3-27B      & 1.000 & 1.000 & 1.000 & 0.975 \\
\midrule
\multirow{7}{*}{Gold Set}
 & GPT-4-Turbo      & 0.983 & 0.983 & 1.000 & 1.000 \\
  & GPT-4-32k        & 1.000 & 0.983 & 1.000 & 0.980 \\
 & Gemma-3-27B      & 0.967 & 1.000 & 0.983 & 0.967 \\
  & Phi-4            & 0.917 & 0.933 & 1.000 & 0.950 \\
 & OpenChat-3.5     & 0.783 & 0.750 & 0.900 & 0.783 \\  
 & LLaMA-3-8B       & 0.683 & 0.683 & 0.817 & 0.733 \\
 & Mistral-7B       & 0.617 & 0.650 & 0.850 & 0.650 \\
\midrule
\multirow{7}{*}{Synthetic Set}
 & GPT-4-32k        & 1.000 & 1.000 & 1.000 & 1.000 \\
 & GPT-4-Turbo      & 1.000 & 1.000 & 1.000 & 1.000 \\
 & Gemma-3-27B      & 0.906 & 1.000 & 1.000 & 0.875 \\
  & Phi-4            & 0.750 & 0.750 & 0.800 & 0.775 \\
 & LLaMA-3-8B       & 0.700 & 0.900 & 0.950 & 0.700 \\
 & Mistral-7B       & 0.688 & 0.844 & 0.906 & 0.688 \\
 & OpenChat-3.5     & 0.594 & 0.812 & 0.844 & 0.688 \\
\bottomrule
\end{tabular}
\caption{Explanation quality scores by dataset and model across four dimensions.}
\label{tab:explanation_quality}
\end{table}
%%%%%%%%

%%%%%%%%CONFIDENCE
% \subsection{Confidence and Coverage (Detailed)}
% Table~\ref{tab:appendix-confidence-coverage}
\begin{table}[H]
\centering
\scriptsize
\setlength{\tabcolsep}{3pt}
\begin{tabular}{lccccc}
\toprule
\textbf{Model} & \textbf{Gold Conf.} & \textbf{Gold Cov. \%} & \textbf{Synth. Conf.} & \textbf{Synth. Cov. \%} & \textbf{Eff. Conf.} \\
\midrule
\texttt{GPT-4-Turbo}     & 4.174 & 100.0 & 4.639 & 100.0 & 4.64 \\
\texttt{GPT-4-32k}       & 4.283 & 100.0 & 4.440 & 100.0 & 4.44 \\
\texttt{Phi-4}           & 4.464 & 100.0 & 4.711 & 99.5  & 4.69 \\
\texttt{Gemma-3-27B}     & 4.265 & 100.0 & 4.698 & 47.6  & 2.24 \\
\texttt{OpenChat-3.5}    & 4.204 & 99.9  & 4.249 & 47.4  & 2.01 \\
\texttt{Mistral-7B}      & 4.237 & 98.9  & 4.132 & 47.6  & 1.97 \\
\texttt{LLaMA-3-8B}      & 4.311 & 92.9  & 3.981 & 37.6  & 1.50 \\
\bottomrule
\end{tabular}
\caption{Average model confidence and coverage across Gold and Synthetic Sets.}
\label{tab:appendix-confidence-coverage}
\end{table}

%%%%%%%%

% \end{onecolumn}
\twocolumn

\end{document}